\documentclass{article} % For LaTeX2e
\usepackage{iclr2017_conference,times}
\usepackage{hyperref}
\usepackage{url}

\usepackage{algorithm}
\usepackage{algorithmic}
\usepackage{bbm}
\usepackage{bm}
\usepackage{color}
\usepackage{graphicx}
\usepackage{amsthm}
\usepackage{amsmath}
\usepackage{amssymb}
\usepackage{caption}
\usepackage{subcaption}
\usepackage{float}
\usepackage{wrapfig}
\usepackage{adjustbox}
\usepackage{enumitem}
\usepackage{lipsum}
\usepackage{multirow}
\usepackage{defs_151115}
\usepackage{hhline}

\title{Generative Adversarial Parallelization}

\author{Daniel Jiwoong Im \\
AIFounded Inc.\\
Toronto, ON\\
\texttt{\{daniel.im\}@aifounded.com} \\
\And
He Ma, Chris Dongjoo Kim and Graham W.~Taylor\\
University of Guelph\\
Guelph, ON\\
\texttt{\{hma02,ckim07,gwtaylor\}@uoguelph.ca} 
}

\newtheorem*{definition}{Definition}
\DeclareMathOperator*{\argmax}{argmax}

\begin{document}

\maketitle

\begin{abstract}
  Generative Adversarial Networks (GAN) have become one of the most
  studied frameworks for unsupervised learning due to their intuitive
  formulation. They have also been shown to be capable of generating
  convincing examples in limited domains, such as low-resolution
  images. However, they still prove difficult to train in practice and
  tend to ignore modes of the data generating
  distribution. Quantitatively capturing effects such as mode coverage
  and more generally the quality of the generative model still remain
  elusive. We propose Generative Adversarial Parallelization (GAP), a
  framework in which many GANs or their variants are trained
  simultaneously, exchanging their discriminators. This eliminates the
  tight coupling between a generator and discriminator, leading to
  improved convergence and improved coverage of modes. We also propose
  an improved variant of the recently proposed Generative Adversarial
  Metric and show how it can score individual GANs or their
  collections under the GAP model.
\end{abstract}

\section{Introduction}

The growing popularity Generative Adversarial Networks (GAN) and
their variants stems from their success in producing realistic samples
\citep{Denton2015, Radford2015, Im2016gran, Salimans2016,
  Dumoulin2016} as well as the intuitive nature of the adversarial
training framework \citep{Goodfellow2014}.  Compared to other
unsupervised learning paradigms, GANs have several merits:
\begin{itemize}[leftmargin=*]
    \item The objective function is not restricted to distances in input
      (e.g.~pixel) space, for example, reconstruction error. Moreover,
      there is no restriction to certain type of functional forms such as
      having a Bernoulli or Gaussian output distribution.
    %\item Compared to variational inference-based models 
    %    \citep{Kingma2014vae, Mnih2014, Gregor2015} 
    %    there is no need to carefully design a
    %    suitable approximate posterior and derive the corresponding lower
    %    bound on the likelihood.
    \item Compared to undirected probabilistic graphical models
    \citep{Hinton06, Salakhutdinov2009}, samples are
    generated in a single pass rather than iteratively. Moreover, the
    time to generate a sample is much less than recurrent models
    like PixelRNN \citep{Oord2016}.
    \item Unlike inverse transformation sampling models,
    the latent variable size is not restricted  
    \citep{Hyvarinen1999, Dinh2014}.
\end{itemize}

\begin{figure}[t]
    \begin{minipage}{0.105\textwidth}
        \includegraphics[width=\linewidth]{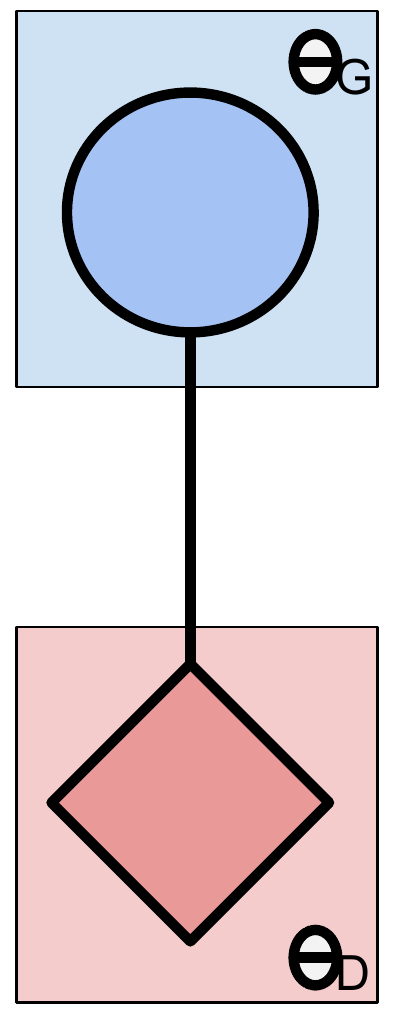}
        \vspace{-0.2cm}
        \subcaption{GAN}
        \label{fig:overview_gan}
    \end{minipage}
    \begin{minipage}{0.49\textwidth}
        \includegraphics[width=\linewidth]{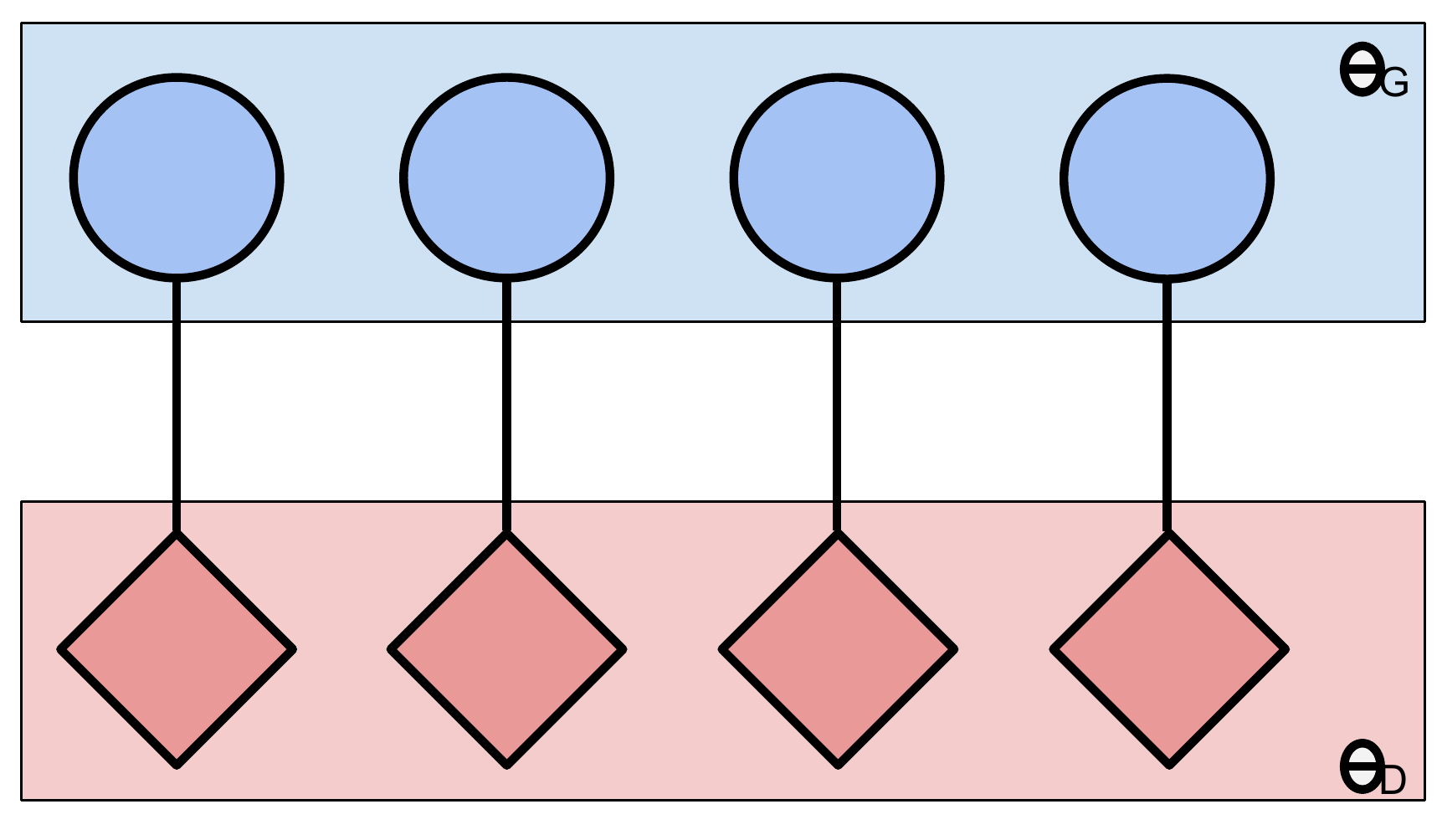}
        \vspace{-0.2cm}
        \subcaption{GANs trained by data-parallelism}
        \label{fig:overview_gan_parallel}
    \end{minipage}
    \begin{minipage}{0.39\textwidth}
        \includegraphics[width=\linewidth]{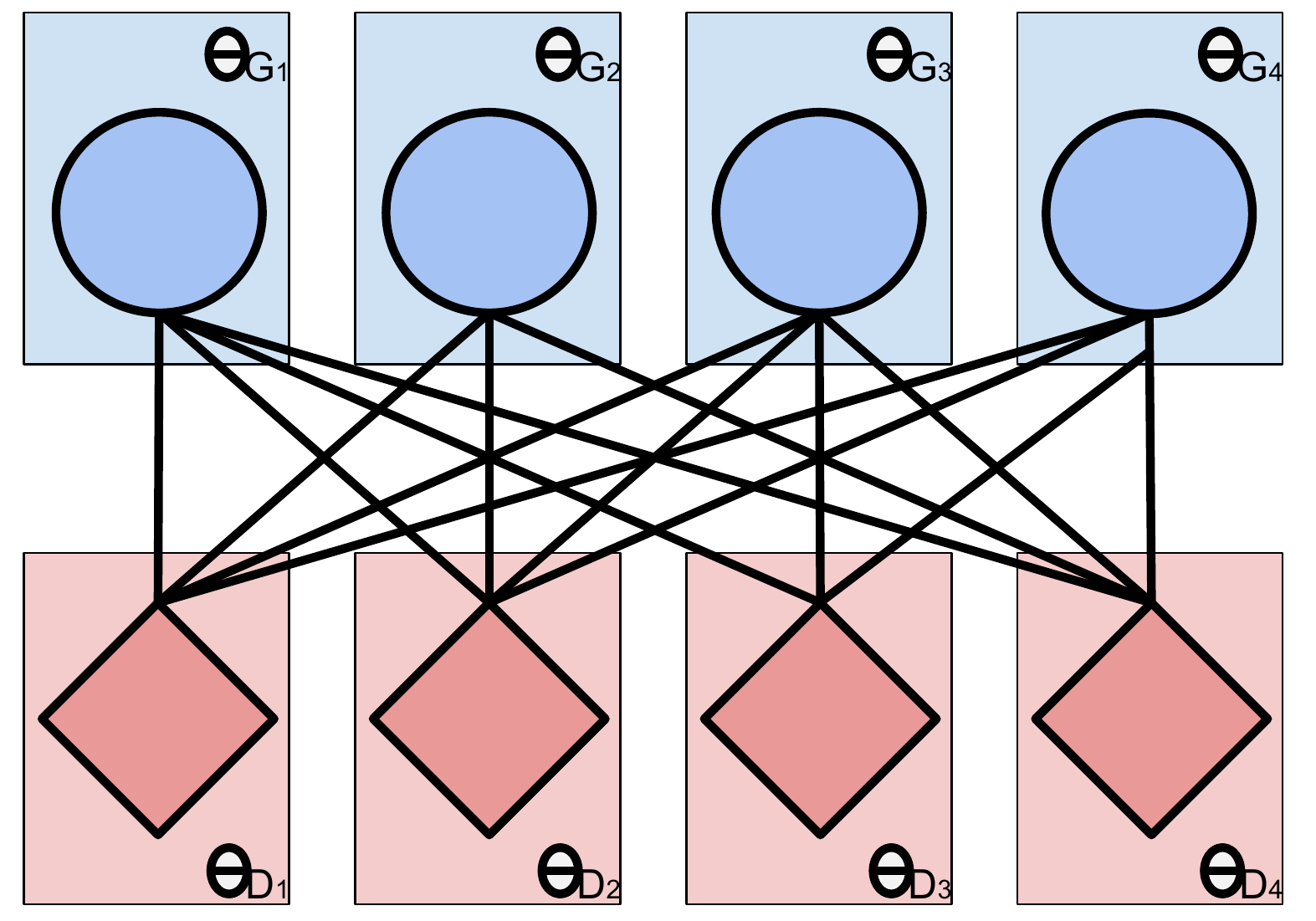}
        \vspace{-0.2cm}
        \subcaption{GAP}
        \label{fig:overview_gan_diag}
    \end{minipage}
    \caption{Depiction of GAN, Parallel GAN, and GAP. Not intended to
      be interpreted as a graphical model.  The difference between
      Figure (b) and (c) is that typical data-based parallelization is
      based on multiple models which share parameters. In contrast,
      GAP requires multiple models with their own parameters which are
      structured in a bipartite formation. }
    \label{fig:overview}
\end{figure}

In contrast, GANs are known to be difficult to train, especially as
the data generating distribution becomes more complex. There have been some attempts to
address this issue.  
For example, \citet{Salimans2016} propose several tricks such as feature
matching and minibatch discrimination.  In this work, we attempt to
address training difficulty in a different way: extending two player generative
adversarial games into a multi-player game. This amounts to training
many GAN-like variants in parallel, periodically swapping their
discriminators such that generator-discriminator coupling is
reduced. Figure \ref{fig:overview} provides a graphical depiction of our method.
% We expound regarding this issue and justify 
% our approach in the next two sections. 

Besides the training dilemma, from the point of view of density
estimation, GANs possess very different
characteristics compared to other probabilistic generative models.
Most probabilistic models
distribute the probability mass over the entire domain, whereas GAN by
nature puts point-wise probability mass near the data. The question of
whether this is desirable property or not is still an open
question\footnote{Noting that the general view in ML is that there is
  nothing wrong with sampling from a degenerate distribution
  \citep{Neal1998}.}.  However, the primary concern of this property is
that GAN may fail to allocate mass to some important modes of the data
generating distribution.  We argue that our proposed model could
alleviate this problem.

That our solution involves training many pairs of generators and
discriminators together is a product of the fact that deep learning
algorithms and distributed systems have been co-evolving for some
time. Hardware accelerators, specifically Graphics Processing Units,
(GPUs) have played a fundamental role in advancing deep learning, in
particular because deep architectures are so well suited to
parallelism \citep{Coates2013-jg}. Data-based parallelism distributes
large datasets over disparate nodes. Model-based parallelism allows
complex models to be split over nodes. In both cases, learning must
account for the coordination and communication among processors. Our
work leverages recent advances along these lines \citep{Ma2016-ed}.

% \begin{itemize}
%     \item data parallelization 
%     \item model parallelization 
% \end{itemize}

%\input{relatedwork}
\section{Background}

The concept of a {\em two player zero-sum game} is borrowed from {\em
  game theory} in order to train a generative adversarial network
\citep{Goodfellow2014}. A GAN consists of a generator $G$ and
discriminator $D$, both parameterized as feed-forward neural networks.
The goal of the generator is to generate samples that fool the
discriminator from thinking that those samples are from the data
distribution $p(\bm x)$, \emph{ad interim} the discriminative
network's goal is to not get tricked by the generator.

This view is formalized into a {\em minimax objective} such that
the discriminator maximizes the expectation of its predictions
while the generator minimizes the expectation of the
discriminator's predictions,
\begin{equation}
    \min_{\bm \theta_G} \max_{\bm \theta_D} V(D,G) 
    = \min_{\bm \theta_G} \max_{\bm \theta_D} \Big[ \mathbb{E}_{\bm x 
        \sim p_{\mathcal{D}}}\big[\log D(\bm x)\big] 
        + \mathbb{E}_{\bm z \sim p_{\mathcal{G}}}\big[\log 
        \big(1-D(G(\bm z))\big)\big] \Big].
    \label{eqn:gan_obj}
\end{equation}
where $\bm \theta_G$ and $\bm \theta_D$ are the parameters
(weights) of the neural networks, $p_{\mathcal{D}}$ is the
data distribution, and $p_{\mathcal{G}}$ is the prior 
distribution of the generative network.

Proposition 2 in \citep{Goodfellow2014} illustrates the ideal concept
of the solution. For two player game, each network's gain of the
utility (loss of the cost) ought to balance out the gain (loss) of the
other network. In this scenario, the generator's distribution becomes
the data distribution. Remark that when the objective function is
convex, gradient-based training is guaranteed to converge to a saddle point.

\subsection{Empirical observations}
\label{sec:emp}
The reality of training GANs is quite different from the ideal
case due to the following reasons: 
\begin{enumerate}[leftmargin=*]
    \item The discriminative and generative networks are 
    bounded by a finite number of parameters, which limits
    their modeling capacity.
    \item Practically speaking, the second term of the objective function in 
    Equation~\ref{eqn:gan_obj} is a bottleneck early on in
    training, where the discriminator can perfectly distinguish the
    noisy samples coming from the generator. The argument of the log
    saturates and gradient will not flow to the generator.
  \item The GAN objective function is known to be non-convex and it is
    defined over a high-dimensional space. This often results in
    failure of gradient-based training to converge.
\end{enumerate}
The first issue comes from the nature of the modelling problem.
Nevertheless, due to the expressiveness of deep neural networks, they
have been shown empirically to be capable of generating natural images
\citep{Radford2015,Im2016gran} by adopting parameter-efficient
convolutional architectures. The second issue is typically addressed
by inverting the generator's minimization into the maximization
formulation in Equation~\ref{eqn:gan_obj} accordingly,
\begin{equation}
    \min_{\bm \theta_G} \log (1-D(G(\bm z))) 
        \rightarrow \max_{\bm \theta_G} \log (D(G(\bm z))).
\end{equation}
This provides better gradient flow in the earlier stages of training
\citep{Goodfellow2014}.

Although there have been cascades of success in image generation tasks
using advanced GANs \citep{Radford2015, Im2016gran, Salimans2016}, all
of them mention the problem of difficulty in training. For example,
\citet{Radford2015} state that {\em the generator ... collapsing all
  samples to a single point ... is a common failure mode observed in
  GANs}. This scenario can occur when the generator allocates most of
its probability mass to a single sample that the discriminator has
difficulty learning. Empirically, convergence of the learning curve
does not correspond to improved quality of samples coming from the GAN
and vice-versa. %(see Figure~\ref{?}).
This is primarily caused by the
third issue mentioned above. Gradient-based optimization methods are
only guaranteed to converge to a Nash Equilibrium for convex
functions, whereas the loss surface of the neural networks used in
GANs are highly non-convex and there is no
guarantee that a Nash Equilibrium even exists.

\begin{figure}[t]
    \centering
    \includegraphics[width=\columnwidth]{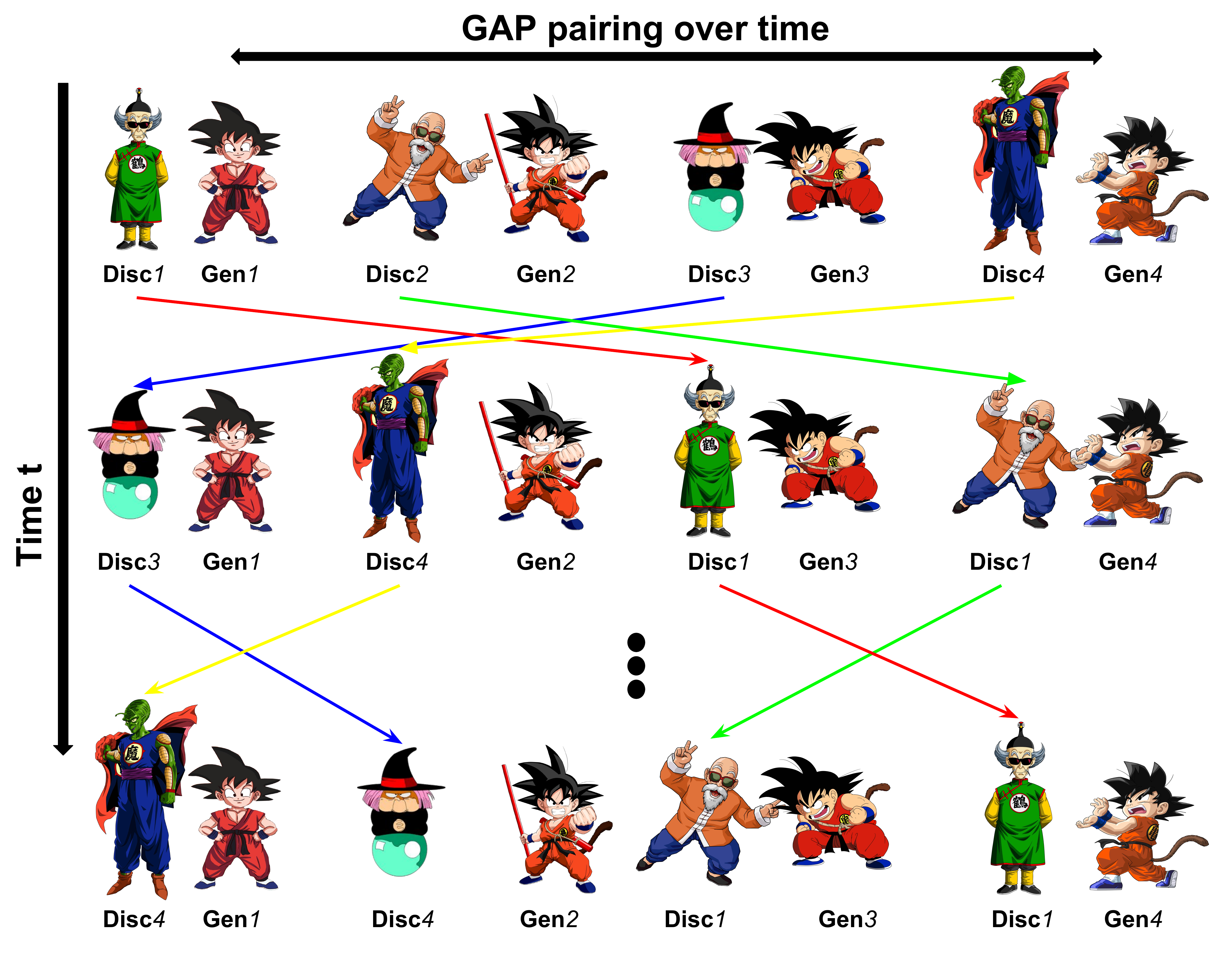}
    \caption{A cartoon illustration of Generative Adversarial Parallelization.
            Generators and discriminators are represented by different monks 
            and sensei.
            The pairing between monks and sensei are randomly substituted 
            overtime.}
    \label{fig:gap_analogy}
\end{figure}

\section{Parallelizing Generative Adversarial Networks}

\begin{wrapfigure}{r}{0.5\textwidth}
%\begin{figure}[htp]
    \vspace{-0.7cm}
    \begin{minipage}{0.5\textwidth}
    \begin{algorithm}[H]
    \begin{algorithmic} 

    \STATE Let $T$ be total number of weight updates.
    \STATE Let $N$ be the total number of GANs.
    \STATE Let $K$ be the swapping frequency.
    \STATE Let $\mathcal{M}=\lbrace (G_1, D_1), (G_2, D_2), \cdots, (G_N,D_N)\rbrace$.

    \WHILE{$t < T$}
        \STATE Update $\mathcal{M}_{i_t}=(G_{i_t}, D_{i_t})$  $\forall i=1\cdots N$.
        \STATE 
        \IF{$t\ \%\ K == 0$}
            \STATE {Randomly select $\frac{N}{2}$ pairs with indices $(i,j)$ w/o replacement.}
            \STATE {Swap $D_i$ and $D_j$ ($G_i$ and $G_j$)  $\forall i \neq j$.}
        \ENDIF
    \ENDWHILE
    \STATE Select the best GAN based on GAM evaluation.
    \end{algorithmic} 
    \caption{Training procedure of GAP.}
    \label{alg:gap}
    \end{algorithm}
    \end{minipage}
    \vspace{-0.5cm}
\end{wrapfigure}

The subject of generative modeling with GANs has undergone intensive
study, and model evaluation between various types of GANs is topic of
increased interest and debate \citep{Theis2015-sm}. Our work is
inspired by the Generative Adversarial Metric \citep{Im2016gran}.
%which directly compares two models $M1:=(G1,D1)$ and $M2:=(G2,D2)$ 
%by having them exchange their discriminators and engage in a “battle” 
%against each other (see the pictorial example in Figure~\ref{fig:gam}).
The GAM enables us to quantitatively evaluate any pair of GANs. The core
concept of the GAM is to swap one discriminator (generator) with the other
discriminator (generator) during the test phase (see the pictorial 
example in Figure~\ref{fig:gam}). 
The GAM concept can easily be extended from evaluation to the training phase.

Our proposal trains multiple GANs simultaneously.  However, unlike the
popular method of data parallelism, we do not train them independently
with shared parameters, rather we try to produce synergy effects among
different GANs during the training phase.  This can be achieved simply
by randomly swapping different discriminators (generators) every $K$
updates.  After training multiple GANS with our proposed method, we
can select the best one based on the GAM.  The pseudocode is shown in
Algorithm~\ref{alg:gap}.

We call our proposed method {\em generative adversarial
  parallelization} (GAP). Note that our method is not model-specific
in a sense that GAP can be applied to any extension of GANs.  For
example, GAP can be applied to DCGAN or GRAN, or we can even apply GAP
on several types of GANs simultaneously.  Say, we have four GPUs
available on which to parallelize models.  We can allocate two GPUs
for DCGANs and the remaining two GPUs for GRANs.  Therefore, we view
GAP as {\em an operator} rather than a model topology/architecture.

\subsection{GAP as regularization}
In a two player generative adversarial game, the concept of
overfitting still exists. However, the realization of overfitting can
be hard to notice. This is mainly due to not having a reconstructive
error function.  For models with a reconstruction-based objective,
samples will simply become identical to the training data as the error
approaches zero. On the other hand, with the GAN objective, even when
the error approaches zero, it does not imply that the samples will
look like the data. So, how can we characterize overfitting in a GAN?

We argue that overfitting in GANs manifests itself differently than in
reconstructive models. Let us explain using two analogies to describe
this phenomenon. Consider a generator as a judo fighter and
discriminator as a sparring partner. When a judo fighter is only
trained with the same sparring partner, his/her fighting strategy will
naturally adapt to the style of his/her sparring partner. Thus, when
the fighter is exposed to a new fighter with a different style, this
fighter may suffer.  Similarly, if a student learns from a single
teacher, his/her learning experience will not only be limited but even
overfitted to the teacher's style of exams (see
Figure~\ref{fig:gap_analogy}).  Equivalently, a paired generator and
discriminator are likely to be adapted to their own strategy.  Here,
GAP intrinsically prevents this problem as the generator
(discriminator) periodically gets paired with different discriminator
(generator). Thus, GAP can be viewed as a regularizer.

\begin{figure}[t]
    \begin{minipage}{0.325\textwidth}
        \includegraphics[width=\linewidth]{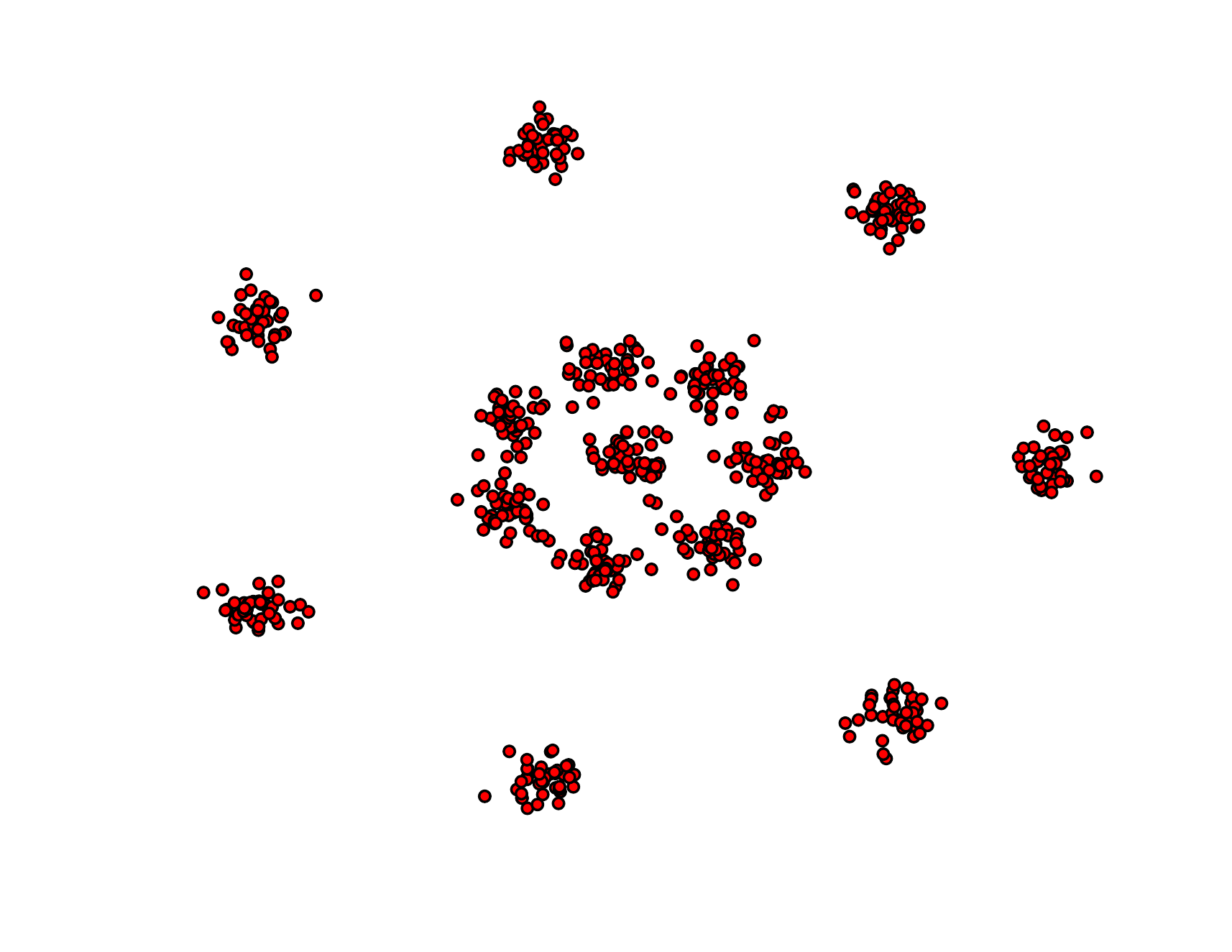}
        \vspace{-0.2cm}
        \subcaption{R15 Dataset}
        \label{fig:syn_data}
    \end{minipage}
    \begin{minipage}{0.325\textwidth}
        \includegraphics[width=\linewidth]{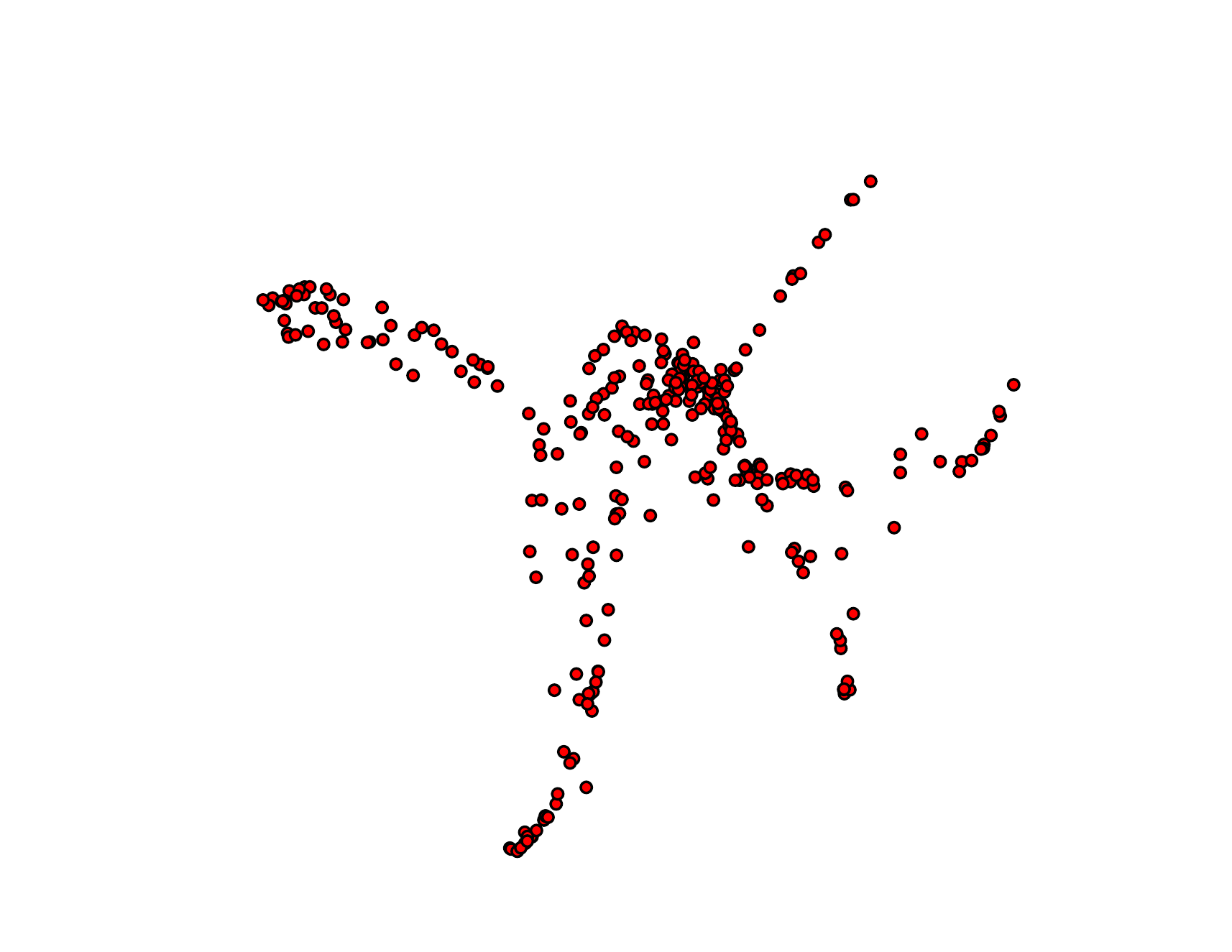}
        \vspace{-0.2cm}
        \subcaption{GAN}
        \label{fig:syn_single}
    \end{minipage}
    \begin{minipage}{0.325\textwidth}
        \includegraphics[width=\linewidth]{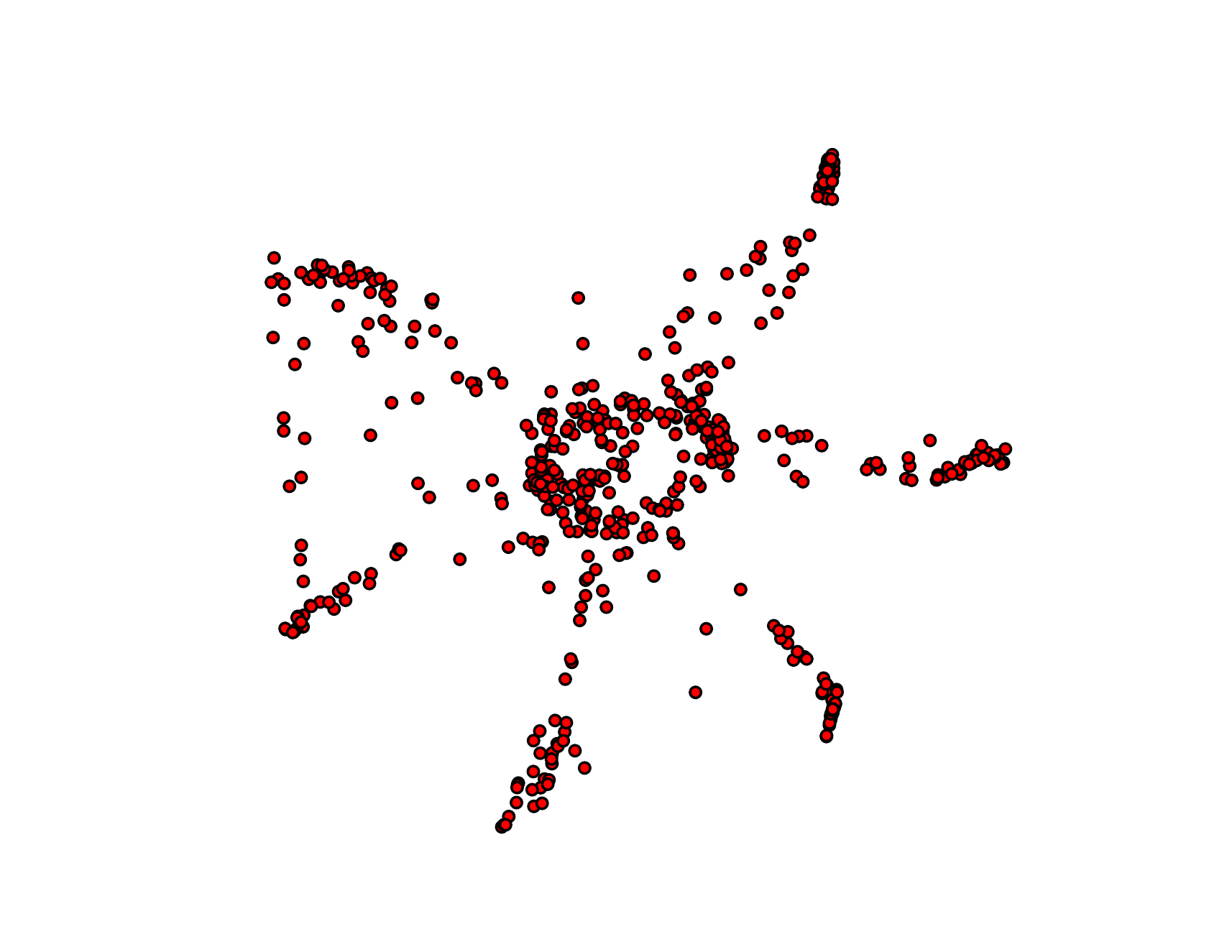}
        \vspace{-0.2cm}
        \subcaption{GAP$_{GAN4}$}
        \label{fig:syn_multi}
    \end{minipage}
    \caption{\subref{fig:syn_data}) The R15 dataset. Samples drawn
      from \subref{fig:syn_single})
             GAN and \subref{fig:syn_multi}) GAP$_{GAN4}$.
             GAP$_{GAP4}$ denotes four GANs trained in parallel 
             with swapping at every epoch.
             The two models were trained using 100 out of 600 data points 
             from the R15 dataset.}
    \label{fig:syn_gen}
    \vspace{0.4cm}
    \begin{minipage}{0.325\textwidth}
        \includegraphics[width=\linewidth]{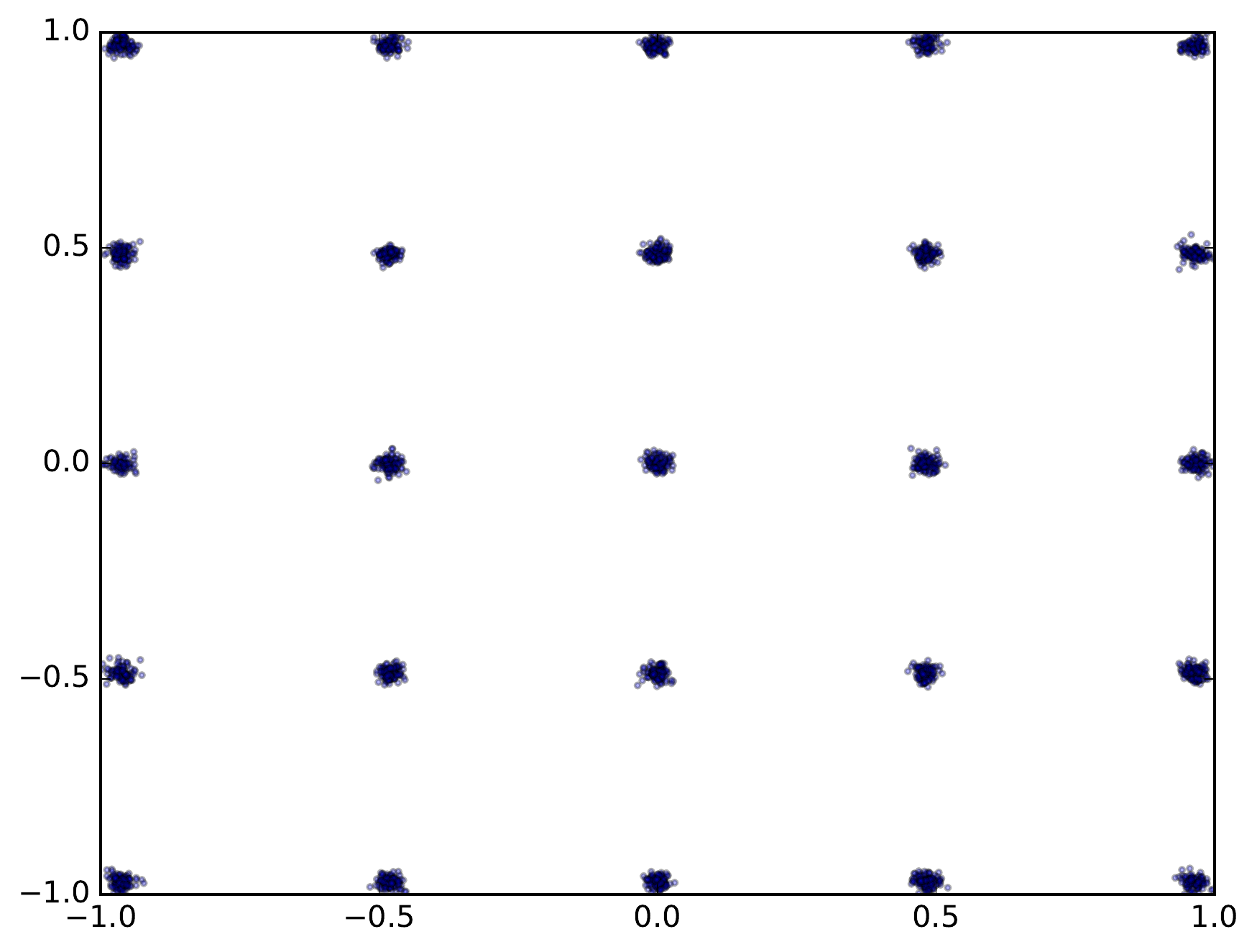}
        \vspace{-0.2cm}
        \subcaption{Mixture of Gaussian}
        \label{fig:mog_data}
    \end{minipage}
    \begin{minipage}{0.325\textwidth}
        \includegraphics[width=\linewidth]{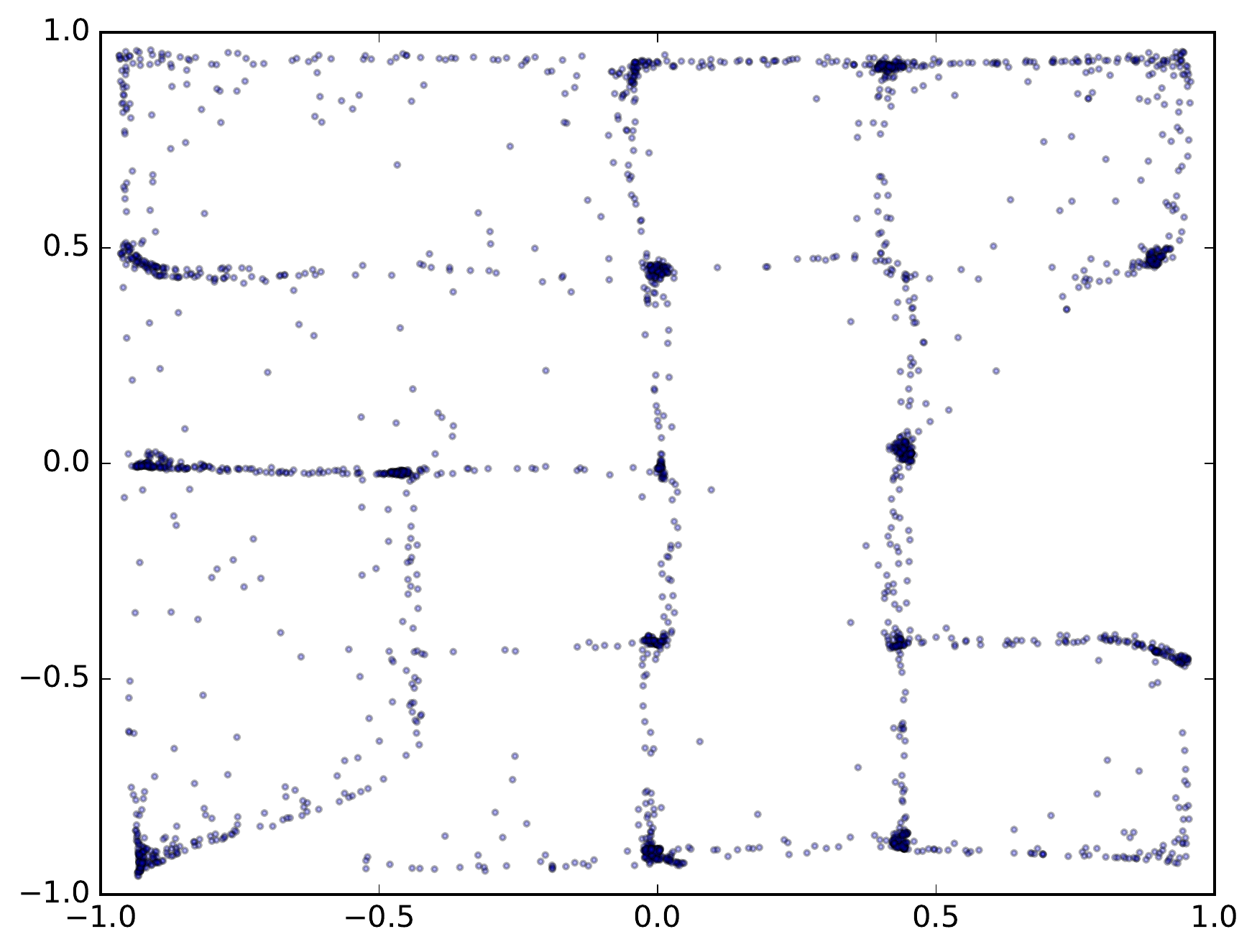}
        \vspace{-0.2cm}
        \subcaption{GAN}
        \label{fig:mog_single}
    \end{minipage}
    \begin{minipage}{0.325\textwidth}
        \includegraphics[width=\linewidth]{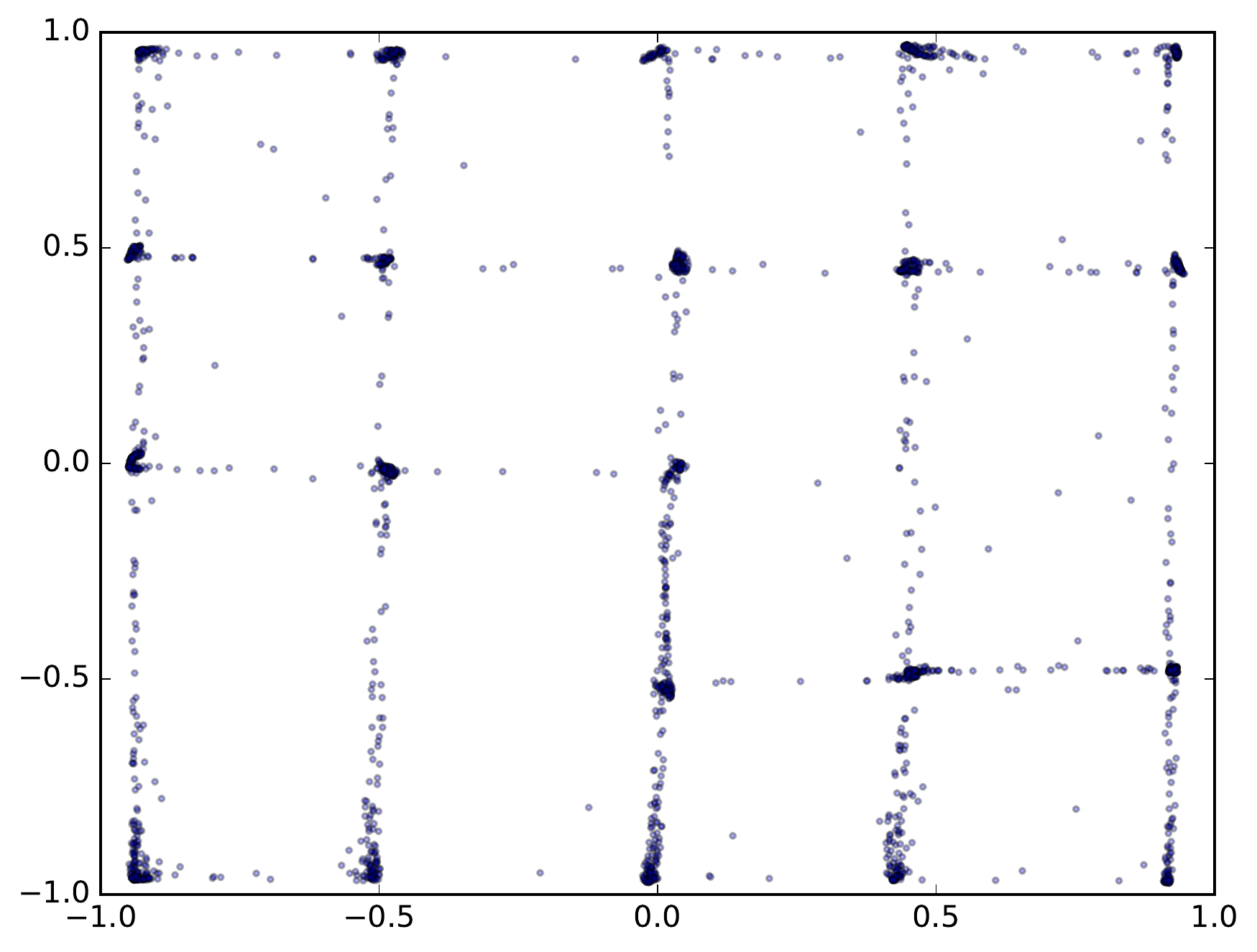}
        \vspace{-0.2cm}
        \subcaption{GAP$_{GAN4}$}
        \label{fig:mog_multi}
    \end{minipage}
    \caption{\subref{fig:mog_data}) The Mixture of Gaussians
      dataset. Samples drawn from \subref{fig:mog_single}) GAN and
      \subref{fig:mog_multi}) GAP$_{GAN4}$. GAP$_{GAP4}$ denotes four
      GANs trained in parallel with swapping at every epoch. The two
      models were trained using 2500 examples.}
    \label{fig:mog_gen}
    \vspace{-0.4cm}
\end{figure}
%\begin{wrapfigure}{r}{0.5\textwidth}
%    \vspace{-1.2cm}
%    \centering
%    \begin{minipage}{0.16\textwidth}
%        \includegraphics[width=\linewidth]{./figs/R15data.pdf}
%        \vspace{-0.2cm}
%        \subcaption{R15}
%        \label{fig:dcgan}
%    \end{minipage}
%    \begin{minipage}{0.16\textwidth}
%        \includegraphics[width=\linewidth]{./figs/gan1_RK100.pdf}
%        \vspace{-0.2cm}
%        \subcaption{DCGAN}
%        \label{fig:dcganx2}
%    \end{minipage}
%    \begin{minipage}{0.16\textwidth}
%        \includegraphics[width=\linewidth]{./figs/gan4_RK100.pdf}
%        \vspace{-0.2cm}
%        \subcaption{GAP$_{DCGAN4}$}
%        \label{fig:dcganx2}
%    \end{minipage}\\
%    \begin{minipage}{0.16\textwidth}
%        \includegraphics[width=\linewidth]{./figs/AggregationData.pdf}
%        \vspace{-0.2cm}
%        \subcaption{Aggregation}
%        \label{fig:dcgan}
%    \end{minipage}
%    \begin{minipage}{0.16\textwidth}
%        \includegraphics[width=\linewidth]{./figs/gan1_Agg.pdf}
%        \vspace{-0.2cm}
%        \subcaption{DCGAN}
%        \label{fig:dcganx2}
%    \end{minipage}
%    \begin{minipage}{0.16\textwidth}
%        \includegraphics[width=\linewidth]{./figs/gan4_Agg.pdf}
%        \vspace{-0.2cm}
%        \subcaption{GAP$_{DCGAN4}$}
%        \label{fig:dcganx2}
%    \end{minipage}\\
%    \caption{DCGAN \& GAP[DCGANx4] are trained on Synthetic dataset, 
%             RK15 and Aggregation, with same hyper-parameters. }
%    \label{fig:gap_cifar10_samples}
%%\end{figure}
%\vspace{-0.7cm}
%\end{wrapfigure}
\subsection{Mode coverage}
The kind of {\em overfitting problem} mentioned above further relates
to the problem of assigning probability mass to different modes of the
data generating distribution -- what we call \emph{mode coverage}.

Let us re-consider the example introduced in Section~\ref{sec:emp}.
Say, the generator was able to figure out a single mode from which samples
are drawn that confuse the discriminator.
As long as the discriminator does not learn to fix 
this problem, the generator is not motivated to consider any other modes.
This kind of scenario allows the generator to cheat by staying within
a single, or small set of modes rather than exploring alternatives.

The story is not exactly the same when there are several different
discriminators interacting with each generator. Since different
discriminators may be good at distinguishing samples from different
modes, each generator must put some effort into fooling all of the
discriminators by generating samples from different modes. The
situation where samples from a single mode fool all of the
discriminators grows much less likely as the number and diversity of
discriminators and generators increases (see
Figure~\ref{fig:syn_gen} and \ref{fig:mog_gen}). Full details of
this visualization are provided in Section \ref{exp_setup}.

\section{Experiments}
\label{sec:experiments}
We conduct an empirical investigation of GAP using two recently
proposed GAN-variants as starting points: DCGAN \citep{Radford2015}
and GRAN \citep{Im2016gran}\footnote{ The Theano-based DCGAN and GRAN implementations
  were based on https://github.com/Newmu/dcgan and
  https://github.com/jiwoongim/GRAN, respectively.}. In each case, we compare
individual GAN-style models to GAP-style ensembles trained in
parallel.

As it is difficult to quantitatively assess mode coverage, first we
aim to visualize samples from GAP vs.~other GAN variants on
low-dimensional (toy) datasets as well as low-dimensional projections
on real data.  Then to evaluate each model quantitatively, we apply
the GAM-II metric which is a re-formulation of GAM \citep{Im2016gran}
which can be used to compare different GAN architectures. Its
motivation and use is described in Section \ref{exp_setup}. 
% GWT: removing this because the following subsection links to the
% Appendix 
% A full description of GAN-II is provided in Appendix~\ref{sec:GAM2}.
We consider, in total, five GAP variants which are summarized in
Table~\ref{tab:model_names}.

\begin{table}[htp]
\caption{GAP variants and their short-hand labels considered in our
  experiments.}
\label{tab:model_names}
\centering
    {\small 
    \begin{tabular}{ | l | c | c | }
      \hline  
      {\bf Name} & {\bf Model} & {\bf Description} \\ 
      \hline\hline
      GAP$_{D2}$ & GAP(DCGAN$\times$2) & Two DCGANs trained with GAP. \\
      GAP$_{D4}$ & GAPDC(DCGAN$\times$4) & Four DCGANs trained with GAP. \\
      GAP$_{G2}$ & GAP(GRAN$\times$2) & Two GRANs trained with GAP. \\
      GAP$_{G4}$ & GAP(GRAN$\times$4) & Four GRANs trained with GAP. \\
      GAP$_{C4}$ & GAP(DCGAN$\times$2, GRAN$\times$2) & 
                  Two DCGANs and two GRANs trained with GAP.\\
        %GAP$_{F4}$ & GAP(DCGAN$\times$2, GRAN$\times$2) & Fine-tuning
        %                                                  with GAP two of each {\em trained} DCGANs and GRANs.\\
    \hline
    \end{tabular}}
\vspace{-0.2cm}
\end{table}

\subsection{Experimental setup}
\label{exp_setup}
%It is worth while to describe the implmentation of GAP.
All of our models are implemented in Theano \citep{Bergstra2010} -- a
Python library that facilitates deep learning research.  Because every
update of each model is implemented as a separate process during
training, swapping their parameters among different GANs necessitates
interprocess communication\footnote{We used openMPI for implementing
  GAP -- see https://www.open-mpi.org/}.  Similar to the Theano-MPI
framework, we chose to do inter-GPU memory transfer instead of passing
through host memory in order to reduce communication overhead.  Random
swapping of the two discriminators' parameters is achieved with an
in-place \verb+MPI_SendRecv+ operation as DCGAN and GRAN share the
same architecture and therefore the same parameterization.

Throughout the experiments, all datasets were normalized between
$[0,1]$.  We used the same hyper-parameters reported in
\citep{Radford2015} and \citep{Im2016gran} for DCGAN and GRAN,
respectively.  The only additional hyper-parameter introduced by GAP
is the frequency of swapping discriminators during training. We also
made deliberate fine-grained distinctions among each GAN trained under
GAP. These were: i) the generator's prior distribution was selected as
either uniform or Gaussian; ii) the order of mini-batches was permuted
during learning; and iii) noise was injected at the input during
learning and the amount of noise was decayed over time. The point of
introducing these distinctions was to avoid multiple GANs converging
to the same or very similar solutions.  Lastly, we used gradient
clipping \citep{Pascanu2013} on both discriminators and generators.

To measure the performance of GANs, our first attempt was to apply GAM
to evaluate our model.  Unfortunately, we realized that GAM is not
applicable when comparing GAP vs.~non-GAP models.  This is because GAM
requires the discriminator the GANs under comparison to have similar
error rates on a held-out test set.  However, as shown in
Figure~\ref{fig:gap_cifar10_learning_curve}, GAP boosts the
generalization of the discriminators, which causes it to have
different test error rates compared to the error rate from non-GAP
models.  Hence, we propose a new metric that omits the GAM's
constraints which we call GAM-II.  It simply measures the average (or
worst case) error rate among a collection of discriminators.
% GWT: I removed this because I didn't feel that it added to clarity
% Note that the best error rate among many
% discriminator is equivalent to looking at the worst performance of the
% generator as it fooled the least.
A detailed description of GAM-II is provided in
Appendix~\ref{sec:GAM2}.

\begin{figure}[htp]
    \centering
    \begin{minipage}{0.495\textwidth}
        \includegraphics[width=\linewidth]{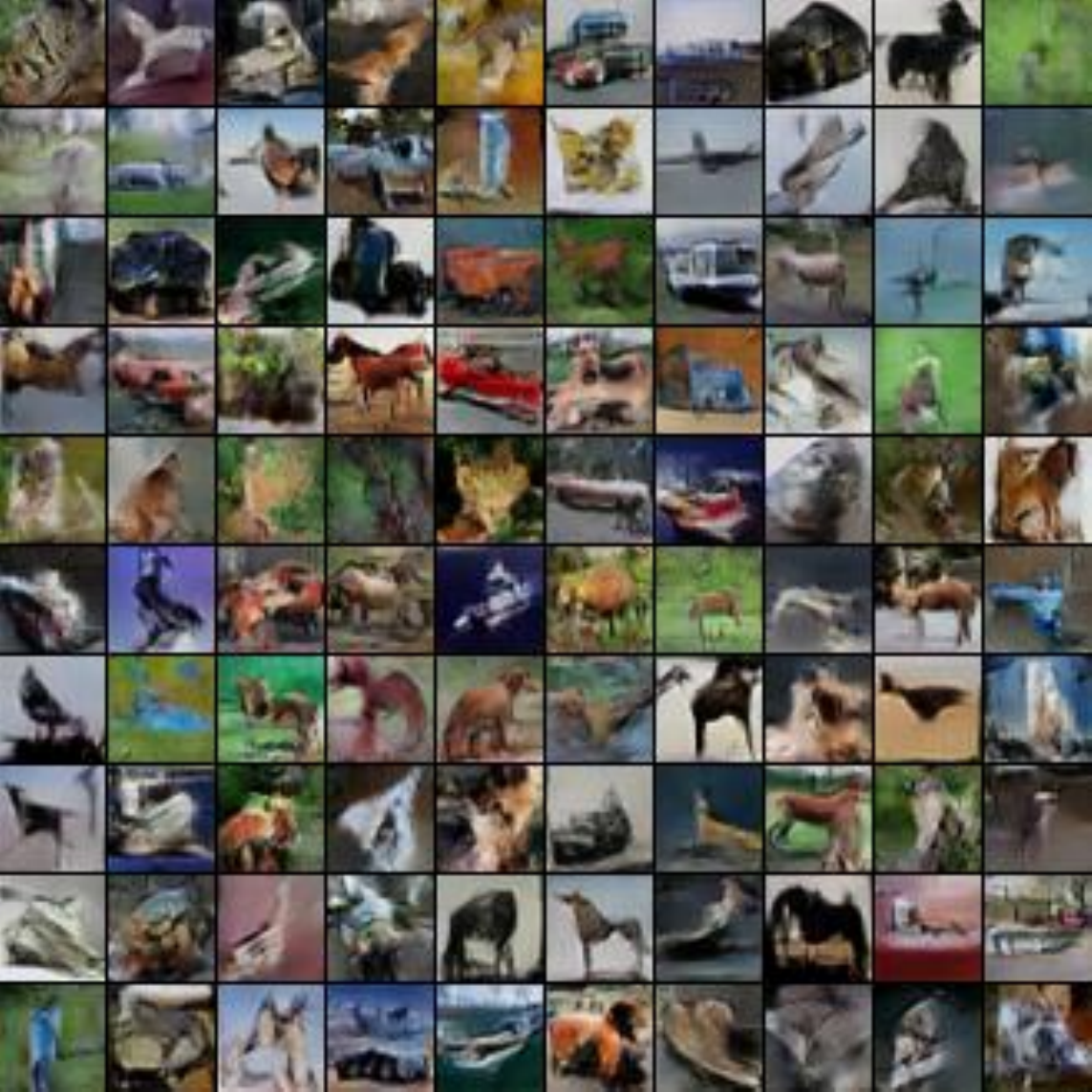}
        \vspace{-0.2cm}
        \subcaption{GAP(DCGANx4)}
        \label{fig:dcganx4}
    \end{minipage}
    \begin{minipage}{0.49\textwidth}
        \includegraphics[width=\linewidth]{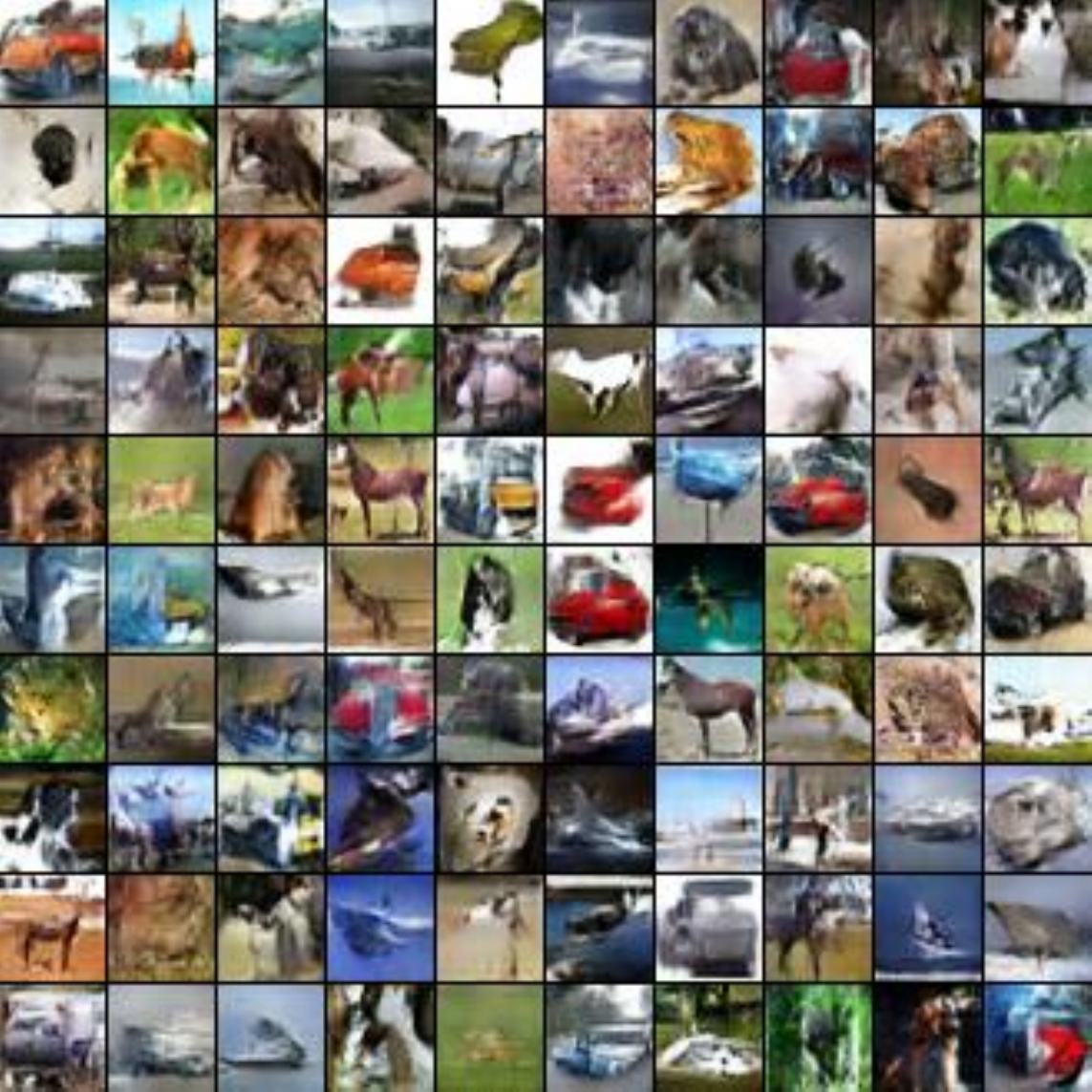}
        \vspace{-0.2cm}
        \subcaption{GAP(GRANx4)}
        \label{fig:granx4}
    \end{minipage}
    \caption{CIFAR-10 samples. Best viewed in colour. More samples 
             are provided in the Appendix.}
    \label{fig:gap_cifar10_samples}
    \vspace{-0.5cm}
\end{figure}

\subsection{Results}

We report our experimental results by answering a few core questions.
% GWT: Redundant with the question
% Let us start by questioning whether GAP covers more modes in synthetic
% datasets.

{\em Q: Do GAP-trained models cover more modes of the data generating distribution?}

Determining whether applying GAP achieves broader mode coverage is
difficult to validate in high-dimensional spaces. Therefore, we
initially verified GAP and non-GAP models on two low-dimensional
synthetic datasets. 
The R15 dataset\footnote{The R15 dataset can be found at
  https://cs.joensuu.fi/sipu/datasets/} contains 500 two-dimensional
data points with 15 clusters as shown in Figure~\ref{fig:syn_data}.
The Mixture of Gaussians dataset\footnote{The Mixture of Gaussians 
dataset can be found at https://github.com/IshmaelBelghazi/ALI/} 
contains 2,500 two-dimensional data points with 25 clusters as shown 
in Figure~\ref{fig:mog_data}.
 
Both discriminator and generator had four fully-connected
batch-normalized layers with ReLU activation units. We first optimized
the hyper-parameters of a single GAN based on visually inspecting the
samples that it generated (i.e.~Figure \ref{fig:syn_gen} shows samples
from the best performing single GAN that we trained). We then trained
four parallelized GANs using the same hyper-parameters of the best
single GAN.

The samples generated from both models are shown in
Figure~\ref{fig:syn_gen} and \ref{fig:mog_gen}.  We observe that
GAP(GAN$\times$4) produces samples that look more similar to the
original dataset compared to a single GAN.  The overlap of samples
generated by four GANs are consistent with Figure~\ref{fig:syn_multi}.
Note that as we decrease the number of training points, the overlap of
GAN samples deviates from the original dataset while GAP seems not to
suffer from this phenomenon.  For example, when we used all 600
examples of R15, both GAN and GAP samples matched the distribution of
data in Figure~\ref{fig:syn_data}. However, as we use less training
examples, GAN failed to accurately model the data distribution by
dropping modes.  The samples plotted in Figure~\ref{fig:syn_multi} are
based on training each model with a random subset of 100 examples drawn from the
original 600.  Based on the synthetic experiments we confirm that GAP
can improve mode coverage when a limited number of training samples
are available.

In order to gain a qualitative sense of models trained using a high
dimensional dataset, we considered two experiments:
i) we examined the class label predictions made on samples from each model
to check how uniformly they were distributed. 
The histogram of the predicted classes is provided in 
Figure~\ref{fig:hist_mnist_dist}. 
ii) we created a t-SNE visualization of generated samples
overlaid on top of the true data (see Appendix~\ref{App:t-SNE}).  We
find that the intersection of data points and samples generated by GAP
is slightly better than samples generated by individual GANs. In
addition to the synthetic data results, these visualizations suggest
some favourable properties of GAP, but we hesitate to draw any strong
conclusions from them.

%Hence, how much GAP helps for mode covering problem in 
%high dimensional space is still an open question.

{\em Q: Does GAP enhance generalization?}

To answer this question, we considered the MNIST, CIFAR-10, and LSUN
church datasets which are often used to evaluate GAN variants.  MNIST
and CIFAR-10 consist of 50,000 training and 10,000 test images of size
27$\times$28 and 32$\times$32$\times$3 pixels, respectively. Each
contains 10 different classes of objects.  The LSUN church dataset
contains various outdoor church images. These high resolution images
were downsampled to 64$\times$64 pixels. The training set consists of
126,227 examples.

One implicit but imperfect way to measure the generalization of a GAN
is to observe generalization of the discriminator alone.  This is because
the generator is influenced by the discriminator and vice versa.  If
the discriminator is overfitting the training data, then the generator
must be biased towards the training data as well.  Here, we plot the
learning curve of the discriminator during training for both
GAP(DCGAN) and GAP(GRAN).

\begin{figure}[h]
    \centering
    \begin{minipage}{0.495\textwidth}
        \includegraphics[width=\linewidth]{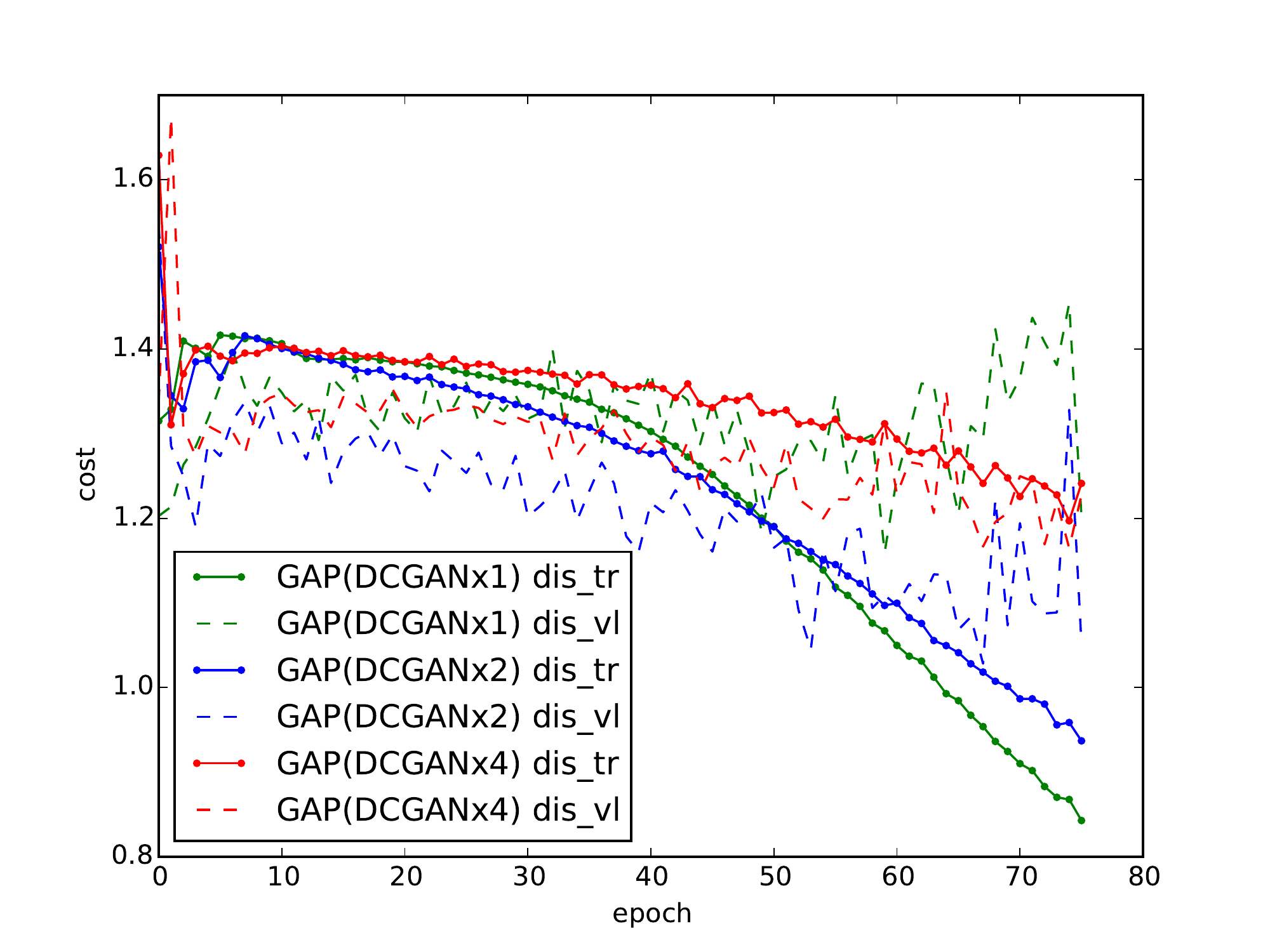}
        \vspace{-0.4cm}
        \subcaption{DCGAN}
        \label{fig:dcganx4}
    \end{minipage}
    \begin{minipage}{0.495\textwidth}
        \includegraphics[width=\linewidth]{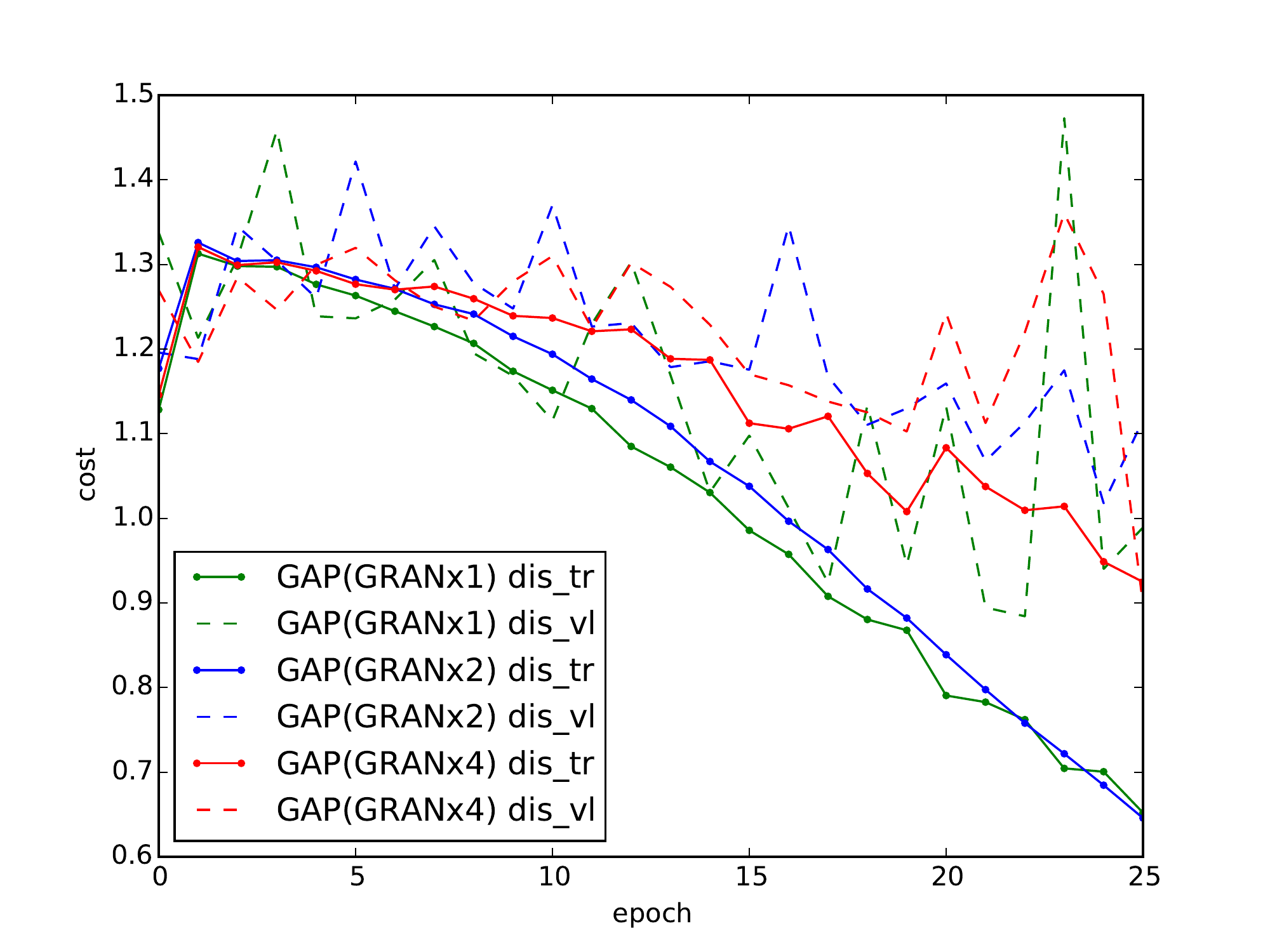}
        \vspace{-0.4cm}
        \subcaption{GRAN}
        \label{fig:granx4}
    \end{minipage}
    \vspace{-0.2cm}
    \caption{Discriminator learning curves on CIFAR-10 as a proxy for
      generalization performance. As parallelization scales up, the
      spread between training and validation cost shrinks. Note that
      the curves corresponding to ``GAP(DCGANx2)'', ``GAP(DCGANx4)'',
      ``GAP(GRANx2)'' and ``GAP(GRANx4)'' are averages of the
      corresponding GAP models.  See
      Figure~\ref{fig:lc_swap_rate_single} in Appendix
      \ref{app:supporting} for the individual curves before
      averaging.}
    \label{fig:gap_cifar10_learning_curve}
\vspace{-0.2cm}
\end{figure}

Figure~\ref{fig:gap_cifar10_learning_curve} shows the learning curve
for a single model versus groups of two and four models parallelized
under GAP.  We observe that more parallelization leads to less of a
spread between the train and validation curves indicating the ability
of GAP to improve generalization. Note that in order to plot a single
representative learning curve while training multiple models under
GAP, we averaged the learning curves of the individual models. To
demonstrate that our observations are not merely attributable to
smoothing by averaging, we show individual learning curves of the
parallelized GANs (see Figure~\ref{fig:lc_swap_rate_single} in
Appendix \ref{app:supporting}).  From now on, we will work with
GAP$_{D4}$ and GAP$_{G4}$.

{\em Q: How does the rate at which discriminators are swapped affect training?}

\begin{figure}[htp]
    \centering
    \begin{minipage}{0.495\textwidth}
        \includegraphics[width=\linewidth]{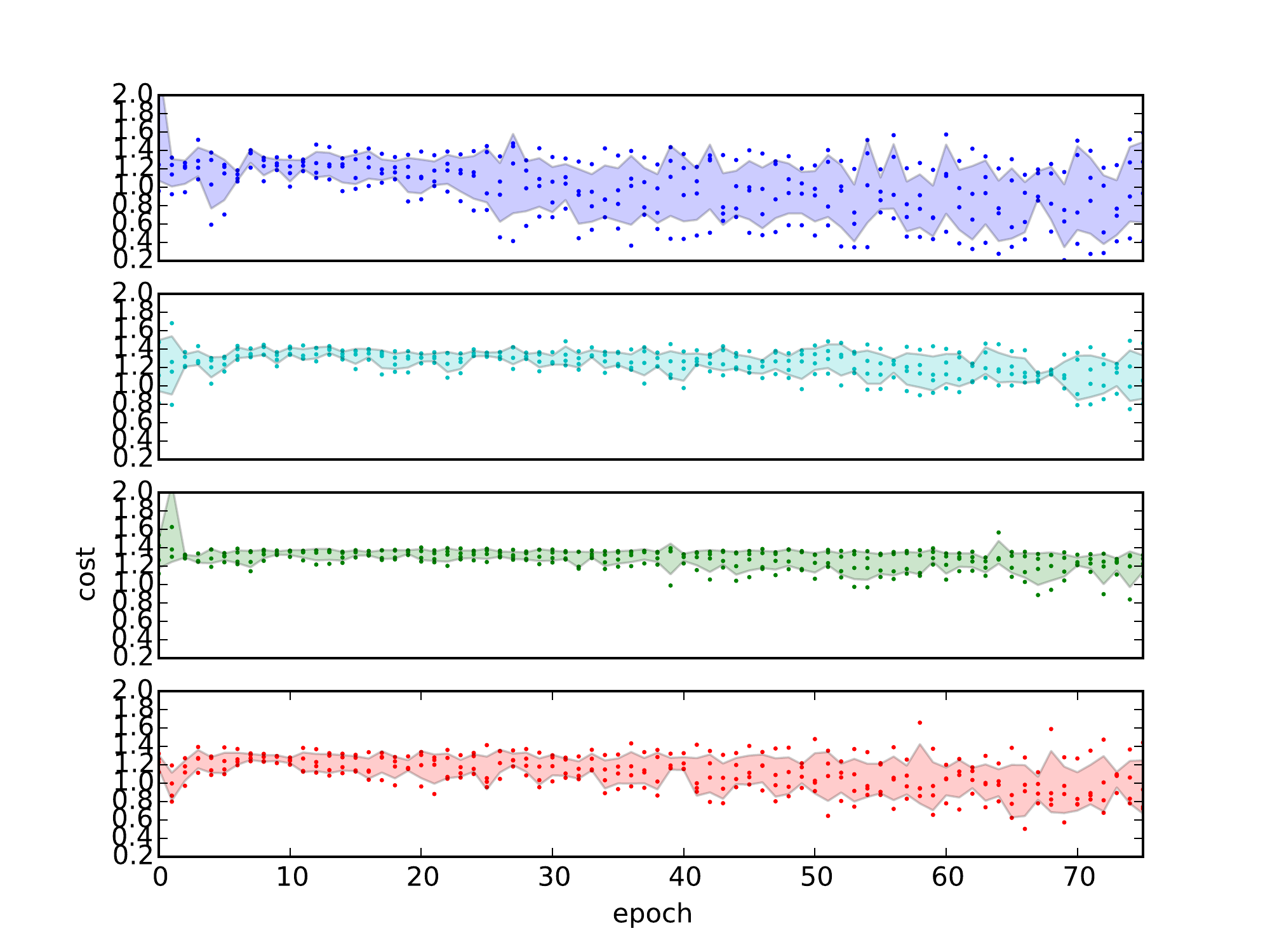}
        \vspace{-0.4cm}
        \subcaption{GAP(DCGAN) trained on CIFAR-10}
        \label{fig:granx4}
    \end{minipage}
    \begin{minipage}{0.495\textwidth}
        \includegraphics[width=\linewidth]{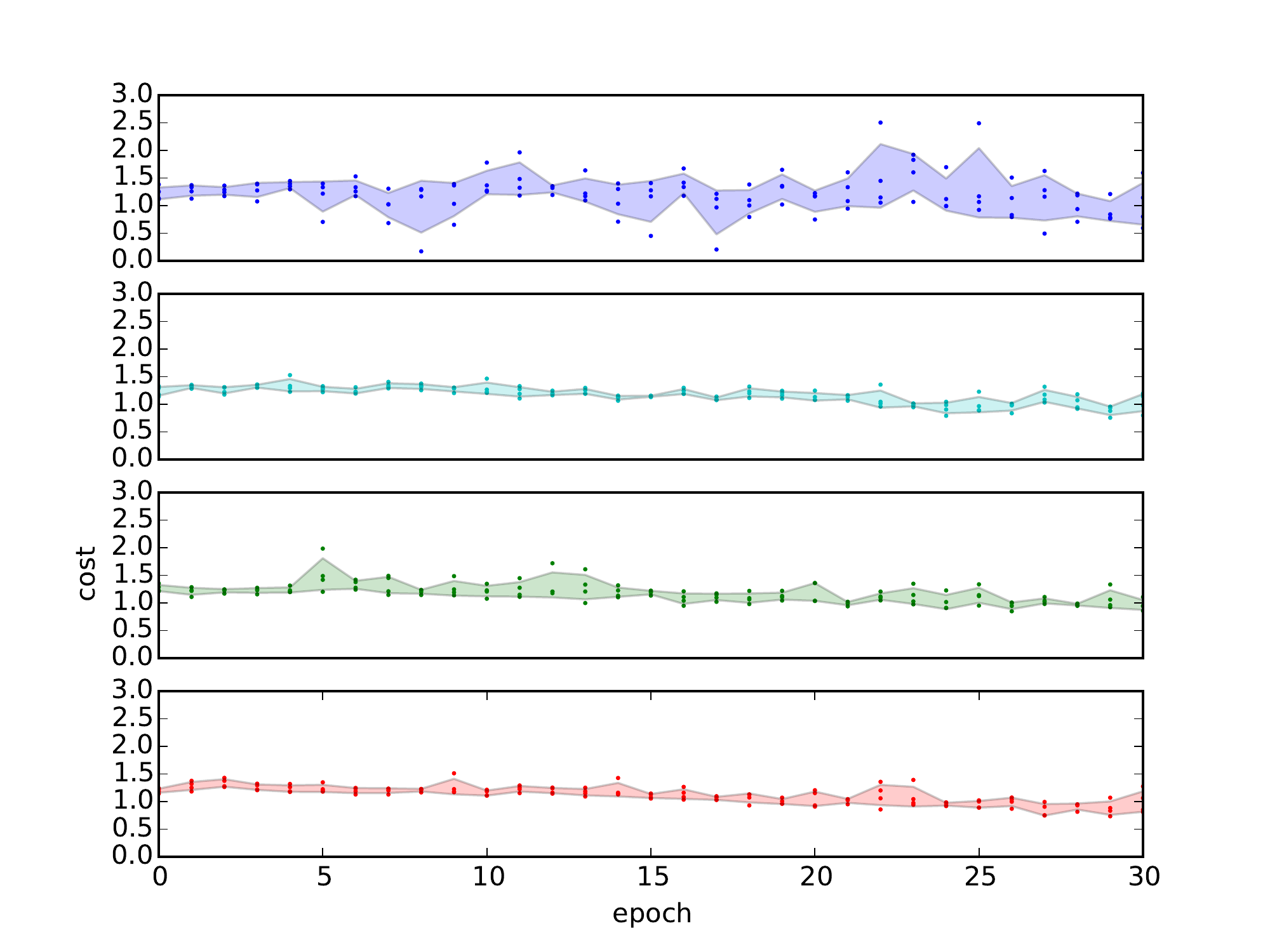}
        \vspace{-0.4cm}
        \subcaption{GAP(GRAN) trained on CIFAR-10}
        \label{fig:granx4}
    \end{minipage}
    \caption{The standard deviations of the validation costs at
      various swapping frequencies. From top to bottom: 0.1, 0.3, 0.5,
      0.7, and 1.0 per epoch.}
    \label{fig:gap_sensitivity_anal}
\vspace{-0.2cm}
\end{figure}
%\vspace{-0.7cm}

As noted earlier, the swapping frequency is the only additional
hyper-parameter introduced by
GAP.  % A sensitivity analysis on swapping frequency is presented in
% Figure~\ref{fig:gap_sensitivity_anal} as well as Figure \ref{fig:gap_lsun_learning_curve}
% in Appendix \ref{app:supporting}.
We conduct a simple sensitivity analysis by plotting the validation
cost of each GAN during training along with its standard deviation in
Figure~\ref{fig:gap_sensitivity_anal}.  We observe that GAP(DCGAN)
varies the least at a swapping frequency of 0.5 -- swapping twice per
epoch.  Meanwhile, GAP(GRANs) are not too sensitive to swapping
frequencies above 0.1. Figure~\ref{fig:gap_lsun_learning_curve} in
Appendix \ref{app:supporting} plots learning curves at different
swapping frequencies. Across all rates, we still see that the spread
between the training and validation costs decreases with the number of
GANs trained in parallel.

%From now on, we will work with swapping frequency of 
%0.5 for both of the models.

{\em Q: Does GAP($\cdot$) improve the quality of generative models?} 

We used GAM-II to evaluate GAP (see Appendix \ref{sec:GAM2}).  We
first looked at the performance over four models: DCGAN, GRAN,
GAP$_{D4}$, and GAP$_{G4}$.  
We also considered combining multiple GAN-variants in a GAP model (hybrid GAP).
We denote this model as GAP$_{C4}$. % and GAP$_{F4}$
GAP$_{C4}$ consists of two DCGANs and two GRANs trained with GAP. % from scratch.
Overall, we have ten generators and ten
discriminators for DCGAN and GRAN: four discriminators from the individually-trained models,
and four discriminators from GAP, and two discriminators from the GAP combination, GAP$_{C4}$.
We used the collection of all ten discriminators to evaluate the
generators.  Table~\ref{tab:GAM2} presents the results.  
Note that we report the minimum and maximum of average and worst error rates 
among four GANs.
Looking at the average errors, GAP$_{D4}$ strongly 
outperforms DCGAN on all datasets.
GAP$_{G4}$ outperforms GRAN on CIFAR-10 and MNIST and strongly
outperforms it on LSUN.
For the case of the maximum worst-case error, GAP
outperforms both DCGAN and GRAN across all datasets.
However, we did not find an improvement on GAP$_{C4}$ based on the GAM-II metric.
%GAP$_{F4}$ consists of two trained DCGANs and two trained GRANs.
%They are then fine-tuned using GAP for five epochs.
%We compared these two models against GAP(DCGAN) and GAP(GRAN)

Additionally, we estimated the log-likelihood assigned by each model
based on a recently proposed evaluation scheme that uses Annealed Importance
Sampling \citep{Wu2016}. 
With the code provided by \citep{Wu2016}, we were able to evaluate
DCGANs trained by GAP$_{D4}$ and
GAP$_{comb}$\footnote{Unfortunately, we did not get the code provided
  by \citep{Wu2016} to work on GRAN.}. 
The results are shown in Table~\ref{tab:ais_eval}. 
Again, these results show that GAP$_{D4}$ improves on DCGAN's performance, but
there is no advantage using combined GAP$_{C4}$.

Samples from each CIFAR-10 and LSUN model for visual inspection are
reproduced in Figures~\ref{fig:gap_cifar10_samples},
\ref{fig:gap_dcgan_lsun10_samples}, \ref{fig:gap_gran_lsun10_samples},
and \ref{fig:comb_gap_cifar10_samples}.
% and \ref{fig:fcomb_gap_cifar10_samples}.

%gap_dcgan_err0 avg: 0.440625, min: 0.2495
%gap_dcgan_err1 avg: 0.4275, min: 0.2255
%gap_dcgan_err2 avg: 0, min: 0
%gap_dcgan_err3 avg: 0.454, min: 0.2645
%dcgan_err0 avg: 0.445875, min: 0
%dcgan_err1 avg 0.411125, min: 0
%dcgan_err2 avg 0.41025, min: 0
%dcgan_err3 avg 0.412, min: 0
%combi_dcgan_err0 avg 0.82525, min: 0
%combi_dcgan_err1 avg 0.8255, min: 0

%gap_gran_err0 avg: 0.297, min: 0.0515
%gap_gran_err1 avg: 0.28075, min: 0.033
%gap_gran_err2 avg: 0.2455, min: 0
%gap_gran_err3 avg: 0.330375, min: 0.0445
%gran_err0 avg: 0.003, min: 0.001
%gran_err1 avg 0.003875, min: 0.001
%gran_err2 avg 0.49725, min: 0.004
%gran_err3 avg 0.000875, min: 0
%combi_gran_err0 avg 0.306125, min: 0
%combi_gran_err1 avg 0.31975, min: 0.023

\begin{table}[htp]
    \centering
    \caption{DCGANs versus GAP(DCGAN) evaluation using GAM-II.}
    \label{tab:GAM2}
    \begin{small}
    \begin{sc}
    \begin{tabular}{c|c|cc|cc|cc}
    \hline
        \multirow{2}{*}{DATASET} & Models   & \multicolumn{2}{c}{DCGAN}  & \multicolumn{2}{c}{GAP$_{D4}$} & \multicolumn{2}{c}{GAP$_{C4}$} \\\hhline{~-------}
                                 & Measure  & Min   & Max    & Min   & Max   & Min   & Max      \\\hline\hline
        \multirow{2}{*}{MNIST}   & Avg.     & 0.352 & 0.395  & 0.430 & 0.476 & 0.398 & 0.423    \\\hhline{~-~~~~}
                                 & Worst    & 0.312 & 0.351  & 0.355 & 0.405 & 0.326 & 0.343    \\\hline
        \multirow{2}{*}{CIFAR-10}& Avg.     & 0.333 & 0.368  & 0.526 & 0.565 & 0.888 & 0.902    \\\hhline{~-~~~~}
                                 & Worst    & 0.173 & 0.225  & 0.174 & 0.325 & 0.551 & 0.615    \\\hline
        \multirow{2}{*}{LSUN}    & Avg.     & 0.592 & 0.628  & 0.619 & 0.652 & 0.108 & 0.180 \\\hhline{~-~~~~}
                                 & Worst    & 0.039 & 0.078  & 0.285 & 0.360 & 0.0   & 0.0 \\
    \hline 
    \end{tabular}
    \end{sc}
    \end{small}
\end{table}

\begin{table}[htp]
    \centering
    \caption{GRAN versus GAP(GRAN) evaluation using GAM-II.}
    \label{tab:GAM2}
    \begin{small}
    \begin{sc}
    \begin{tabular}{c|c|cc|cc|cc}
    \hline
        \multirow{2}{*}{DATASET} & Models   & \multicolumn{2}{c}{GRAN}  & \multicolumn{2}{c}{GAP$_{G4}$} & \multicolumn{2}{c}{GAP$_{C4}$} \\\hhline{~-------}
                                 & Measure  & Min   & Max    & Min   & Max   & Min   & Max      \\\hline\hline
        \multirow{2}{*}{MNIST}   & Avg.     & 0.433 & 0.465  & 0.510 & 0.533 & 0.459 & 0.474    \\\hhline{~-~~~~}
                                 & Worst    & 0.004 & 0.020  & 0.008 & 0.020 & 0.010 & 0.012    \\\hline
        \multirow{2}{*}{CIFAR-10}& Avg.     & 0.289 & 0.355  & 0.332 & 0.416 & 0.306 & 0.319    \\\hhline{~-~~~~}
                                 & Worst    & 0.006 & 0.019  & 0.048 & 0.171 & 0.001 & 0.023    \\\hline
        \multirow{2}{*}{LSUN}    & Avg.     & 0.477 & 0.590  & 0.568 & 0.649 & 0.574 & 0.636 \\\hhline{~-~~~~} 
                                 & Worst    & 0.013 & 0.043  & 0.022 & 0.055 & 0.015 & 0.021 \\                  
    \hline 
    \end{tabular}
    \end{sc}
    \end{small}
\end{table}

\begin{table}[htp]
    \centering
    \caption{The likelihood of DCGANs and GAP(DCGAN) using the AIS
      estimator proposed by \citep{Wu2016} on the MNIST dataset.}
    \label{tab:ais_eval}
    \begin{small}
    \begin{sc}
    \begin{tabular}{c|c|c}
    \hline
        Models   & DCGAN             & GAP$_{D4}$       \\\hline
        AIS      & 682.5 $\pm$ 12.51 & 691.6 $\pm$ 0.01 \\
    \hline 
    \end{tabular}
    \end{sc}
    \end{small}
\end{table}

\section{Discussion}

We have proposed Generative Adversarial Parallelization, a
framework in which several adversarially-trained models are trained
together, exchanging discriminators. We argue that this reduces the
tight coupling between generator and discriminator and show
empirically that this has a beneficial effect on mode coverage,
convergence, and quality of the model under the GAM-II metric. Several
directions of future investigation are possible. This includes
applying GAP to the evolving variety of adversarial models, like
improvedGAN \citep{Salimans2016}. We still view stability as an issue
and partially address it by tricks such as clipping the gradient of
the discriminator. In this work, we only explored synchronous training
of GANs under GAP, however, asynchronous training may provide more
stability. Recent work has explored the connection between GANs and
actor-critic methods in reinforcement learning
\citep{Pfau2016-fs}. Under this view, we believe that GAP may have
interesting implications for multi-agent RL. Although we have assessed
mode coverage qualitatively either directly or indirectly via
projections, quantitatively assessing mode coverage for generative
models is still an open research problem.

% \begin{enumerate}
% \item Apply GAP to other models like improvedGAN. 
% \item Recent work has explored the connection between GANs and RL
%   (actor-critic) (Pau and Vinyals); this has implications for
%   multi-agent RL
% \item Asynchronous training of GAP may stabilize GANs during the
%   training stage; stability is still an issue for us (e.g. clipping
%   the gradient of the discriminator)
% \item Explore GAP with different optimization techniques other than
%         stochastic iterative optimization techniques.
% \item Extention to actor-critic setting in reinforcement learning .
% \item quantitatively assess mode coverage
% \end{enumerate}

\bibliography{main}
\bibliographystyle{iclr2017_conference}

\clearpage
\appendix
\section{Supplementary Material for Generative Adversarial Parallelization}
\begin{figure}[htp]
    \centering
    \includegraphics[width=\columnwidth]{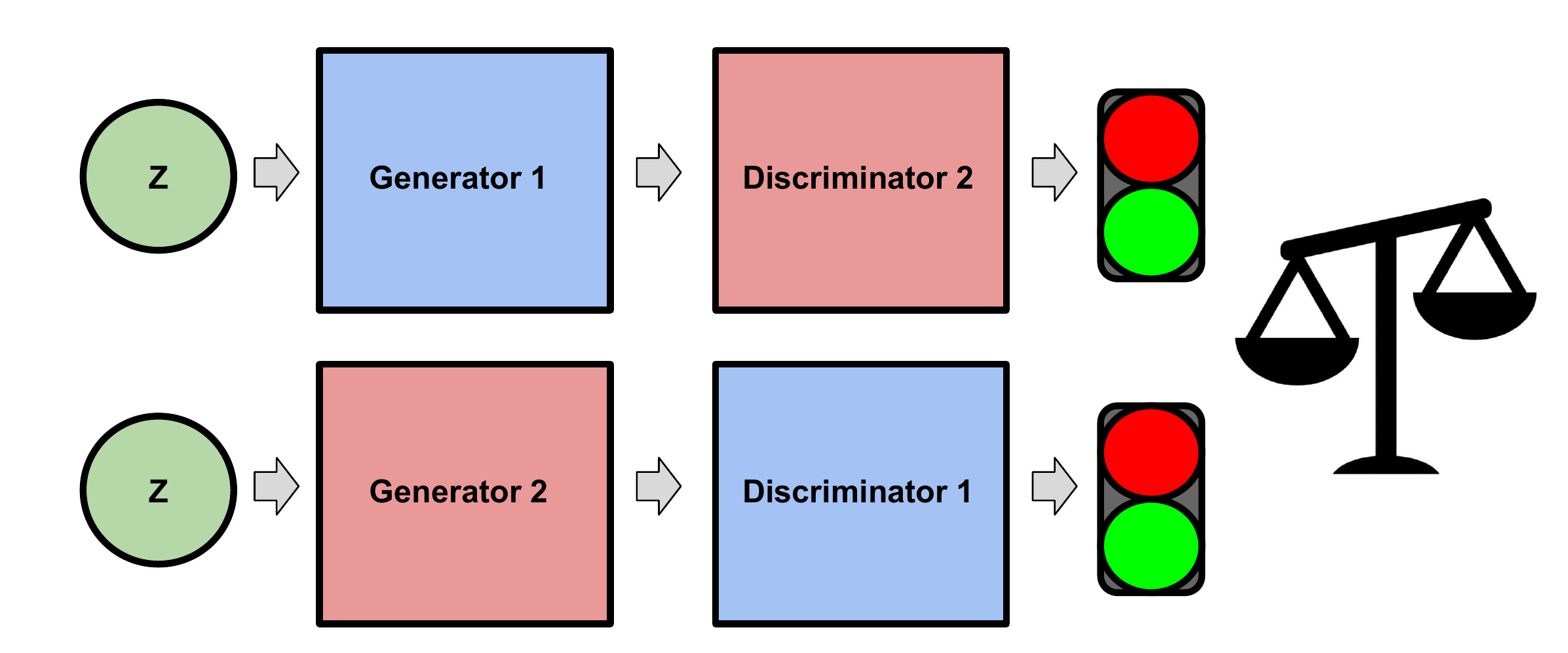}
    \caption{Illustration of the Generative Adversarial Metric.}
    \label{fig:gam}
\end{figure}

\subsection{Generative Adversarial Metric II}
\label{sec:GAM2}
%\begin{wrapfigure}{r}{0.5\textwidth}
\begin{figure}[htp]
    \vspace{-0.2cm}
    \centering
    \includegraphics[width=\linewidth]{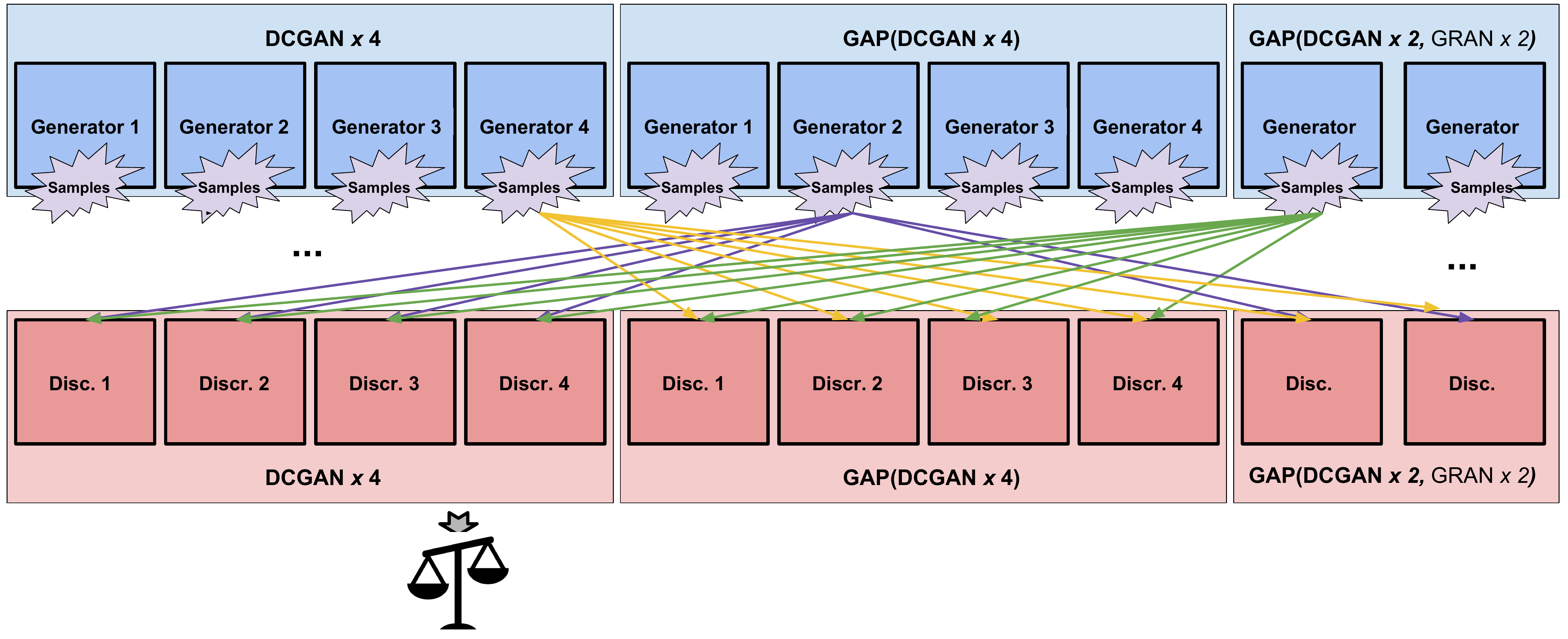}
    \vspace{-0.5cm}
    \caption{Illustration of the Generative Adversarial Metric II.}
    \label{fig:gamII}
    \vspace{-0.2cm}
\end{figure}
%\end{wrapfigure}

Although GAM is a valid metric as it measures {\em the likelihood
  ratio} of two generative models, it is hard to apply in practice.
This is due to having a test ratio constraint, which imposes the
condition that the ratio between test error rates be approximately unity.
However, because GAP improves the generalization of GANs as shown in
Figure~\ref{fig:gap_cifar10_learning_curve}, the test ratio often
does not equal one (see Section \ref{sec:experiments}).  We introduce a new
generative adversarial metric, and call it GAM-II.

GAM-II evaluates a model based on either the average error rate or worst error rate 
of a collection of discriminators given a set of samples from each
model to be evaluated:
\begin{align}
    \argmax_{\lbrace G_j | S_j \sim p_{\mathcal{G}_j} \rbrace} \hat\epsilon(S_j) 
        &= \argmax_{\lbrace G_j | S_j \sim p_{\mathcal{G}_j} \rbrace} \frac{1}{N_j} \sum^{N_j}_{i=1} \epsilon(S_j|D_{i}),\\
    \argmax_{\lbrace G_j | S_j \sim p_{\mathcal{G}_j} \rbrace} \bar \epsilon(S_j) 
        &= \argmax_{\lbrace G_j | S_j \sim p_{\mathcal{G}_j} \rbrace} \min_{i=1\cdots N_j} \epsilon(S_j|D_{i})
\end{align}
where $\epsilon$ outputs the classification error rate, and
$N_j$ is all discriminators except for the ones that the generator $j$
saw during training.
For example, the comparison of DCGAN and GAP applied to four DCGANs is
shown in Figure~\ref{fig:gamII}. 
\begin{definition}
    We say that GAP helps if at least one of the models trained with GAP 
    performs better than a single model. Moreover, GAP strongly-helps
    if all models trained with GAP perform better than a single model.
\end{definition}
In our experiments, we assess GAP based on the definition above.

%Dependencies : openmpi, mpi4py, pycuda, hickle, libgpuarray(pygpu), nccl

%\begin{figure}[htp]
%    \centering
%    \includegraphics[width=\linewidth]{./figs/gap[dcganx2]_large_samples1.pdf}
%    \caption{CIFAR-10 samples generated by GAP[DCGANx2]. (Best viewed in colours) 
%            \imj{Place holder for now. We will update to better 
%                samples as we go.}}
%    \label{fig:gap_cifar10_samples}
%\end{figure}

\subsection{Experiments with t-SNE}

In order to get a qualitative sense of models trained using a high
dimensional dataset, we consider a t-SNE map of generated samples
overlaid on top of the true data.  Normally, a t-SNE map is used to
visualize clusters of embedded high-dimensional data. Here, we are
more interested in the overlap between true data and generated samples
by visualizing clusters which we interpret as modes of the data
generating distribution.

\begin{figure}[htp]
    \centering
    \begin{minipage}{0.495\textwidth}
        \includegraphics[width=\linewidth]{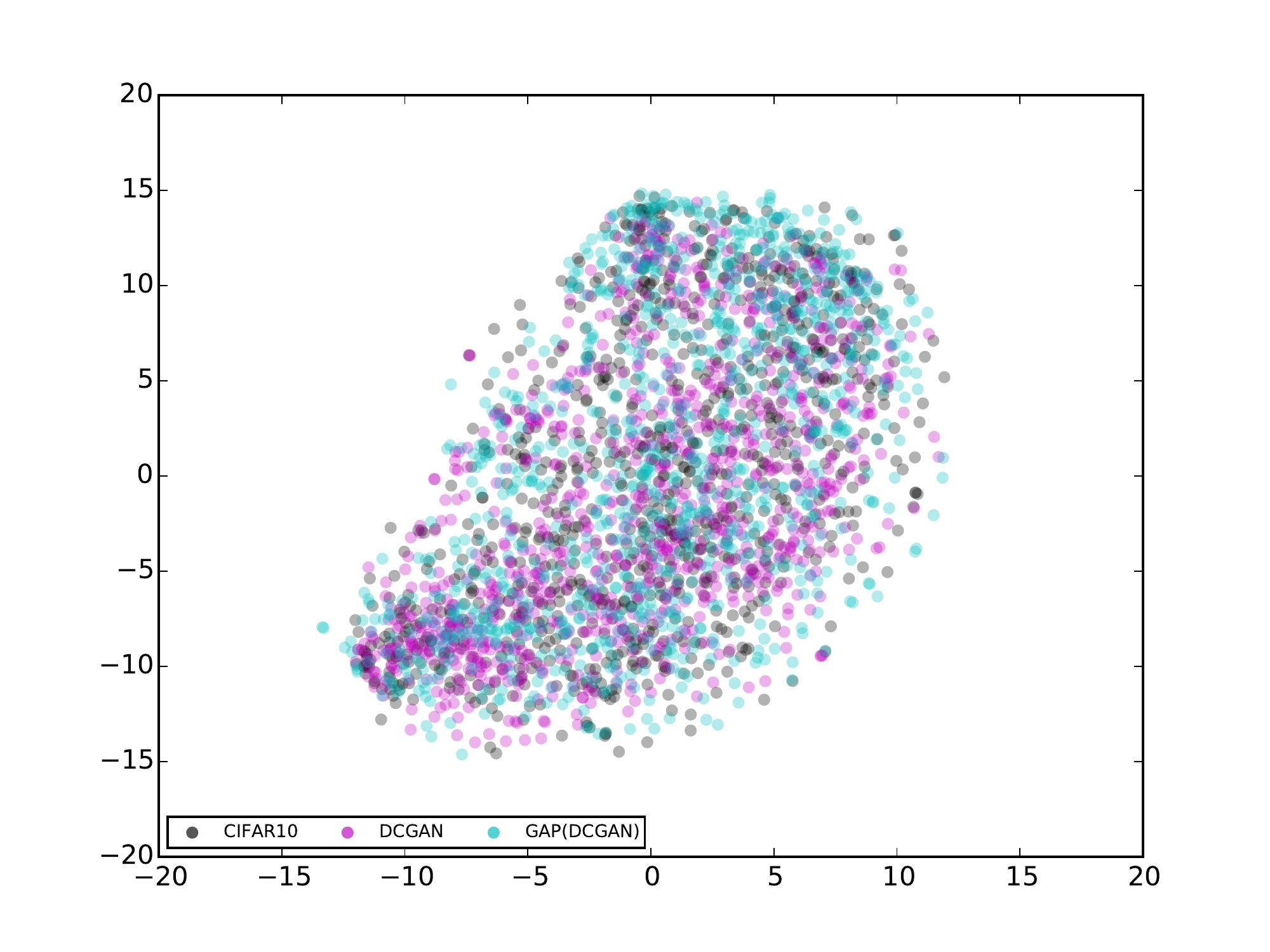}
        \vspace{-0.4cm}
        \subcaption{Using data, DCGAN, \& GAP(DCGAN) samples}
        \label{fig:tsne-dcganx4}
    \end{minipage}
    \begin{minipage}{0.495\textwidth}
        \includegraphics[width=\linewidth]{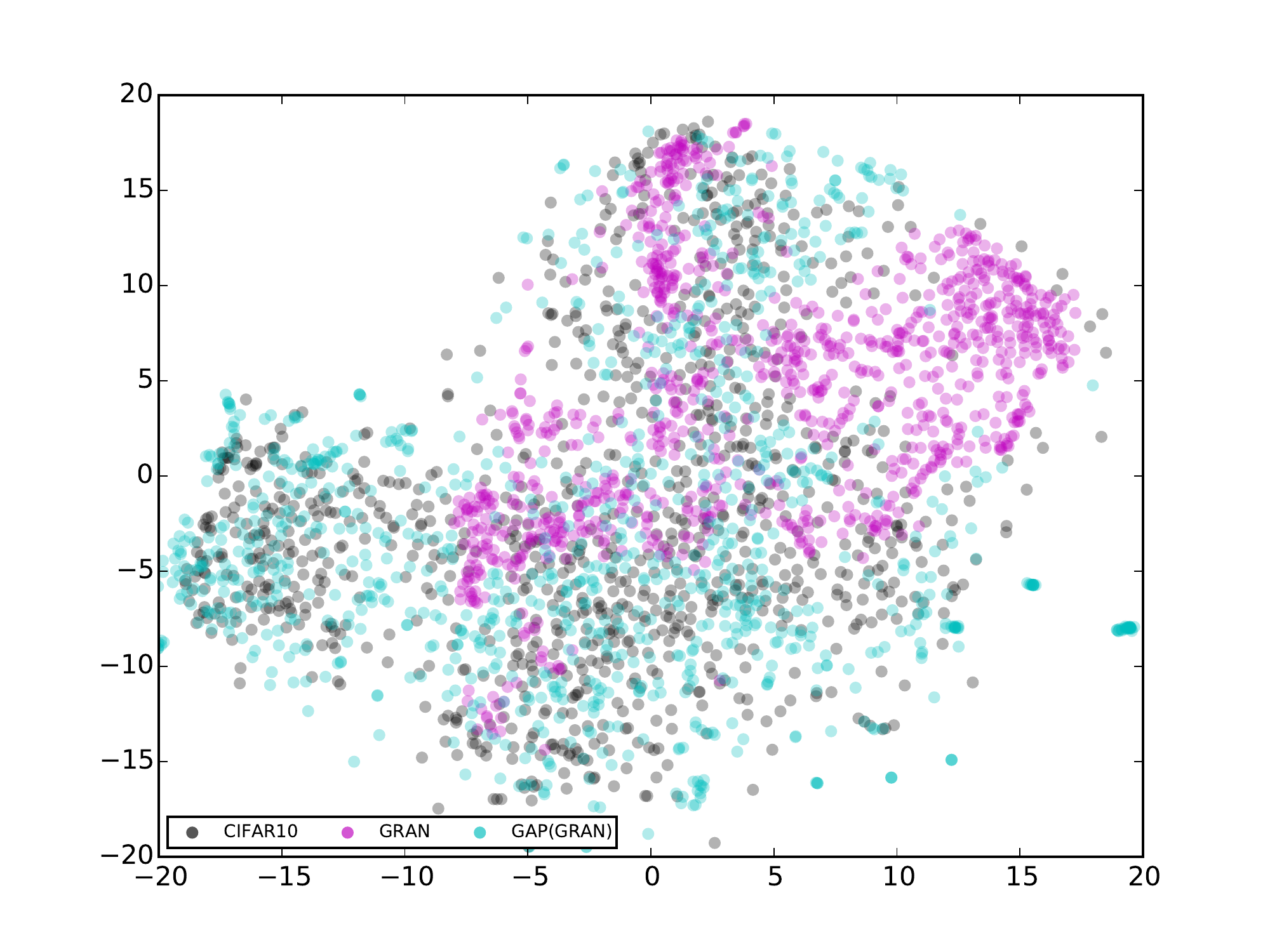}
        \vspace{-0.4cm}
        \subcaption{Using data, GRAN, \& GAP(GRAN) samples}
        \label{fig:tsne-granx4}
    \end{minipage}\\
    \caption{t-SNE mapping of data and sample points on CIFAR-10.  The
      points are colour coded as: Data (Black), Single Model
      (Magenta), and GAP (Cyan). Note that, particularly for the
      figure on the right, there seems to be more overlap between the
      data and the GAP-generated samples compared to the GAN-generated
      samples.}
    \label{fig:gap_tsne_cifar10}
    \begin{minipage}{0.495\textwidth}
        \includegraphics[width=\linewidth]{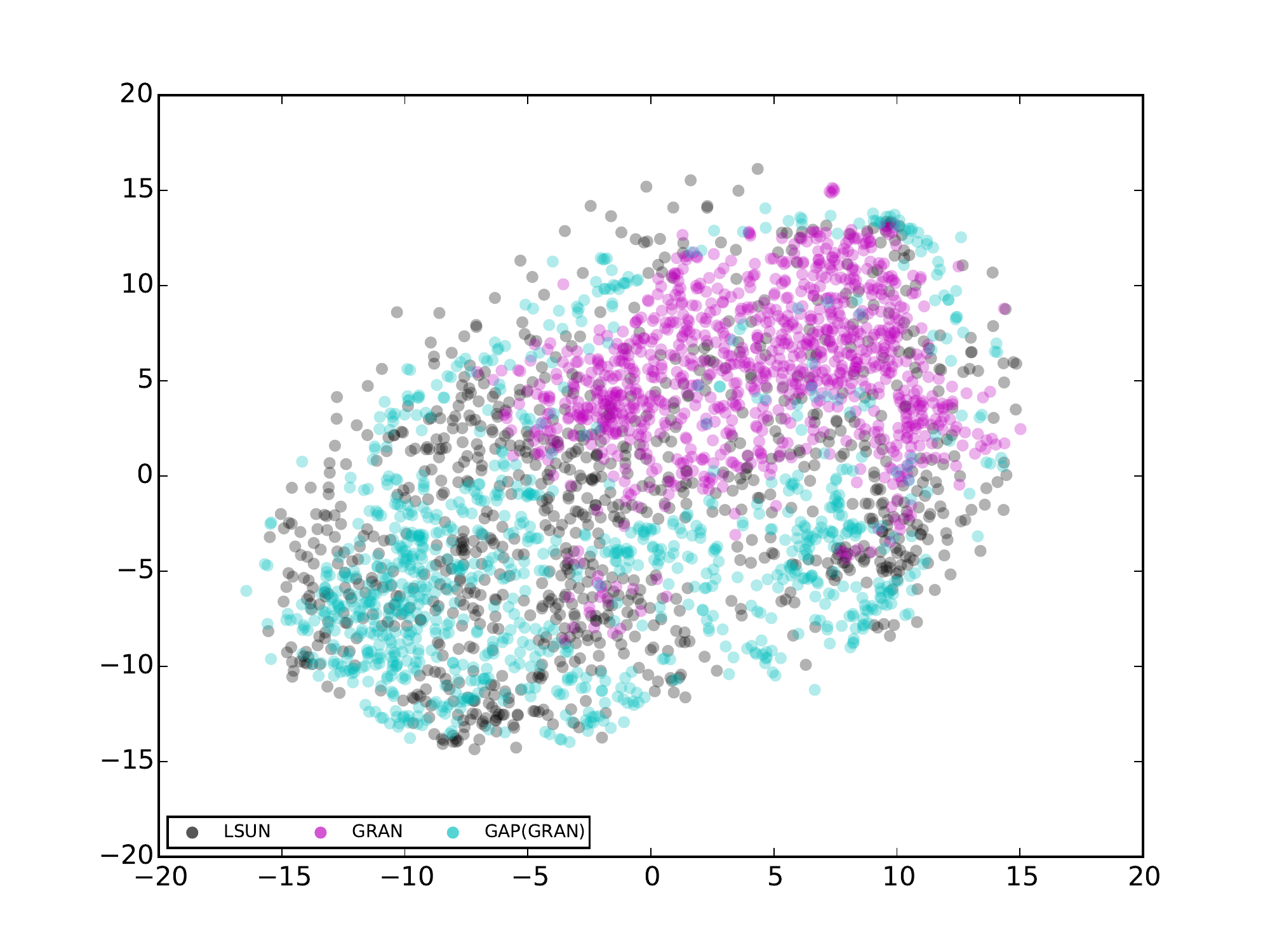}
        \vspace{-0.4cm}
        \subcaption{Using data, DCGAN, \& GAP(DCGAN) samples}
        \label{fig:tsne-dcganx4}
    \end{minipage}
    \begin{minipage}{0.495\textwidth}
        \includegraphics[width=\linewidth]{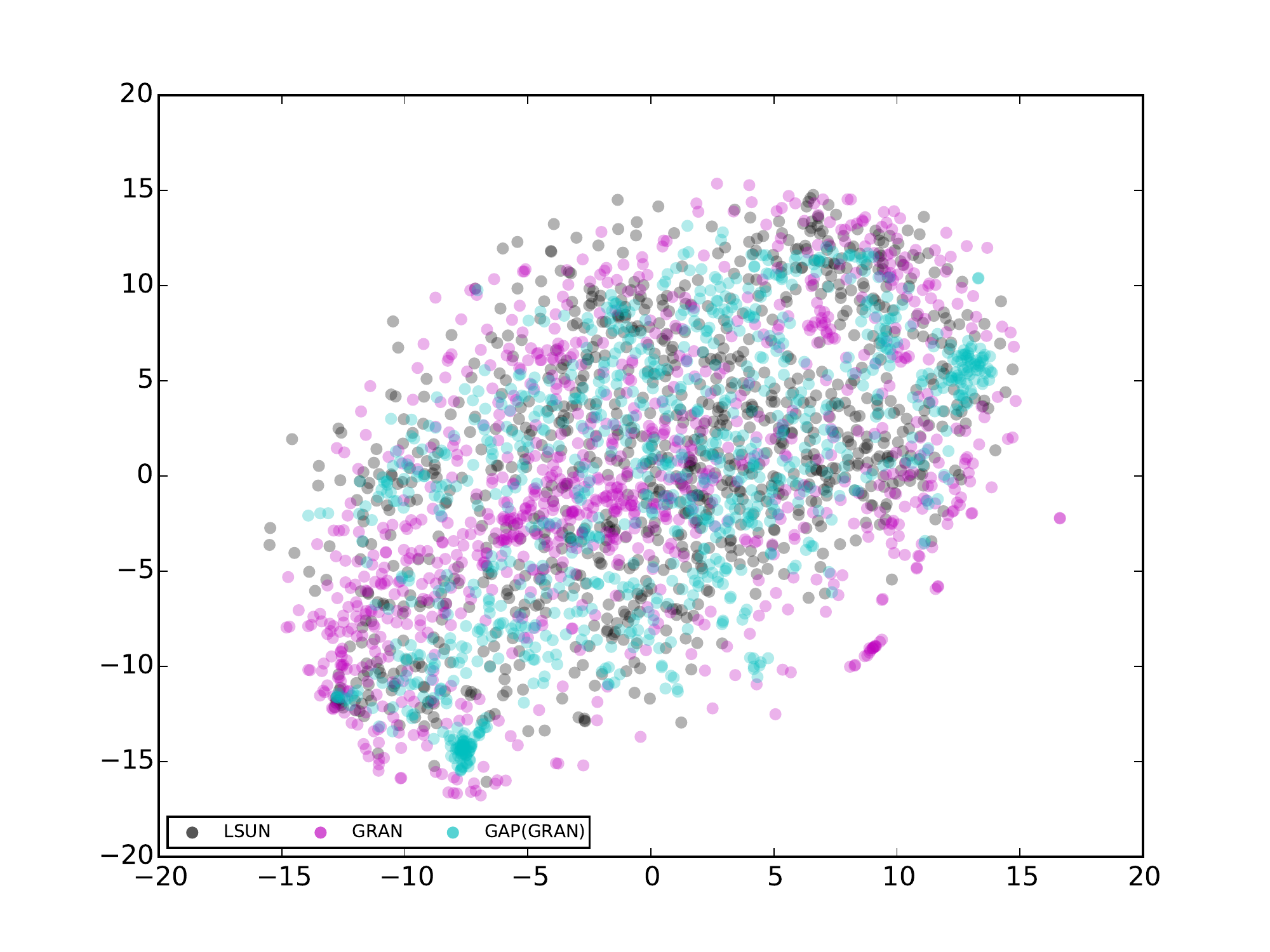}
        \vspace{-0.4cm}
        \subcaption{Using data, GRAN, \& GAP(GRAN) samples}
        \label{fig:tsne-granx4}
    \end{minipage}\\
    \vspace{-0.2cm}
    \caption{t-SNE mapping of data and sample points on the LSUN
      dataset.  The points are colour coded as: Data (Black),
      Single Model (Magenta), and GAP (Cyan).}
    \label{fig:gap_tsne_lsun}
    \vspace{-0.2cm}
\end{figure}

Figure~\ref{fig:gap_tsne_cifar10} and \ref{fig:gap_tsne_lsun} present the
t-SNE map of data and samples from single- and multiple- trained GANs
under GAP.  We find that the intersection of data points and samples
generated by GAP is slightly better than samples generated by
individual GANs. This provides an incomplete view but is nevertheless
a helpful visualization.
\label{App:t-SNE}
\pagebreak

\subsection{Supporting Figures}
\label{app:supporting}
\begin{figure}[htp]
    \begin{minipage}{0.495\textwidth}
        \includegraphics[width=\linewidth]{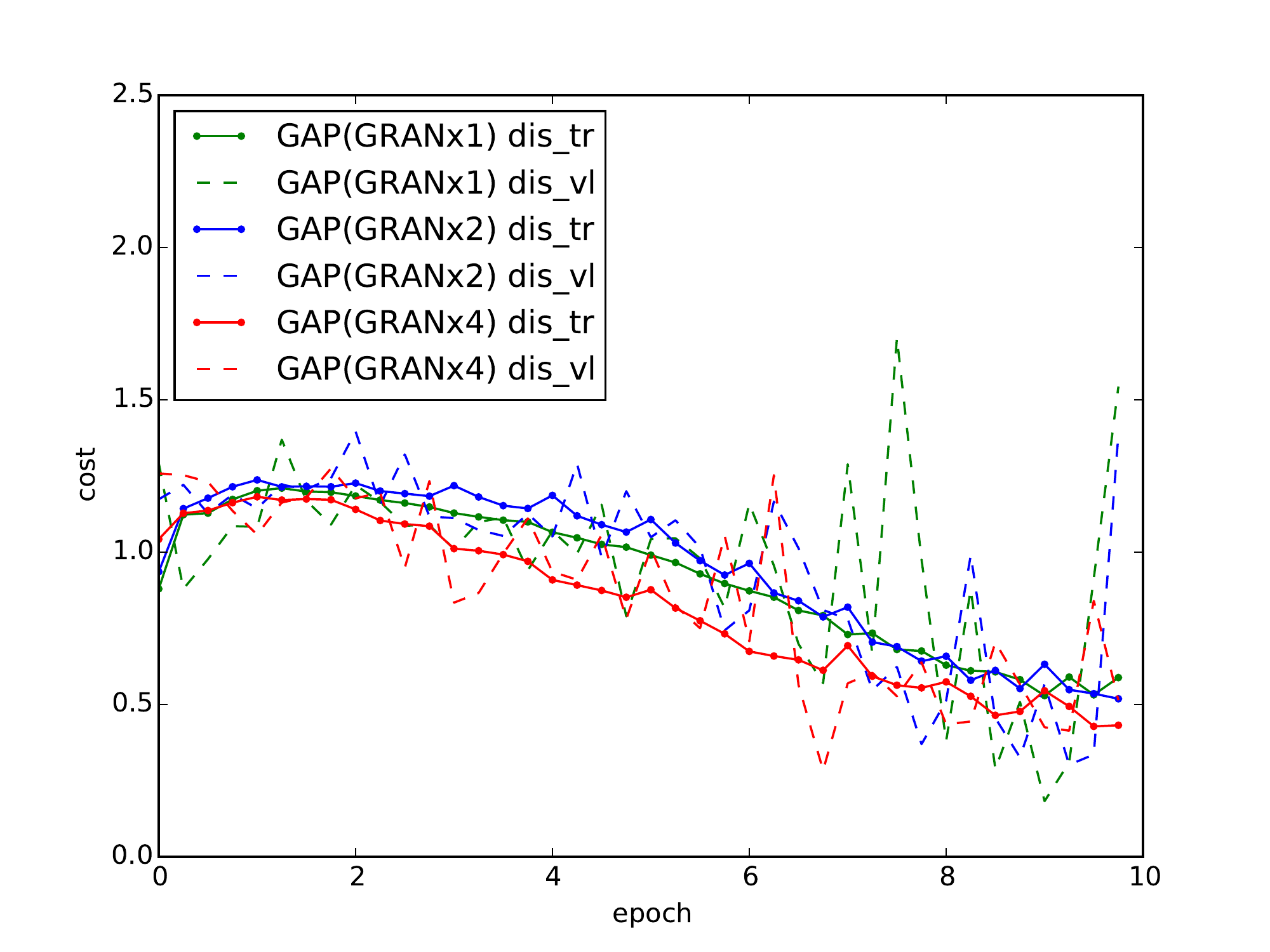}
        \vspace{-0.2cm}
        \subcaption{swapping frequency every 0.1 epoch.}
    \end{minipage}
    \begin{minipage}{0.495\textwidth}
        \includegraphics[width=\linewidth]{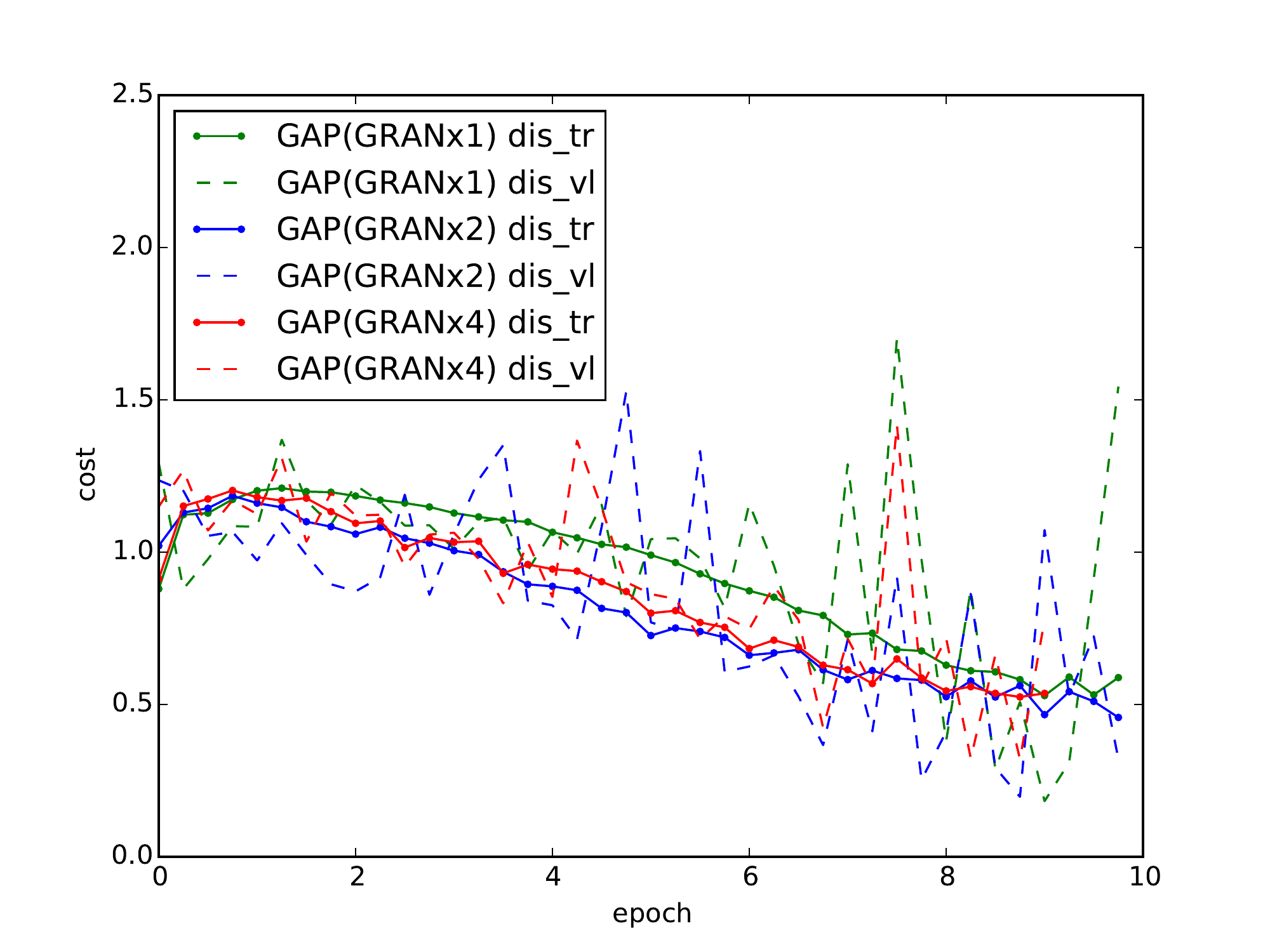}
        \vspace{-0.2cm}
        \subcaption{swapping frequency every 0.3 epoch.}
    \end{minipage}\\
    \begin{minipage}{0.495\textwidth}
        \includegraphics[width=\linewidth]{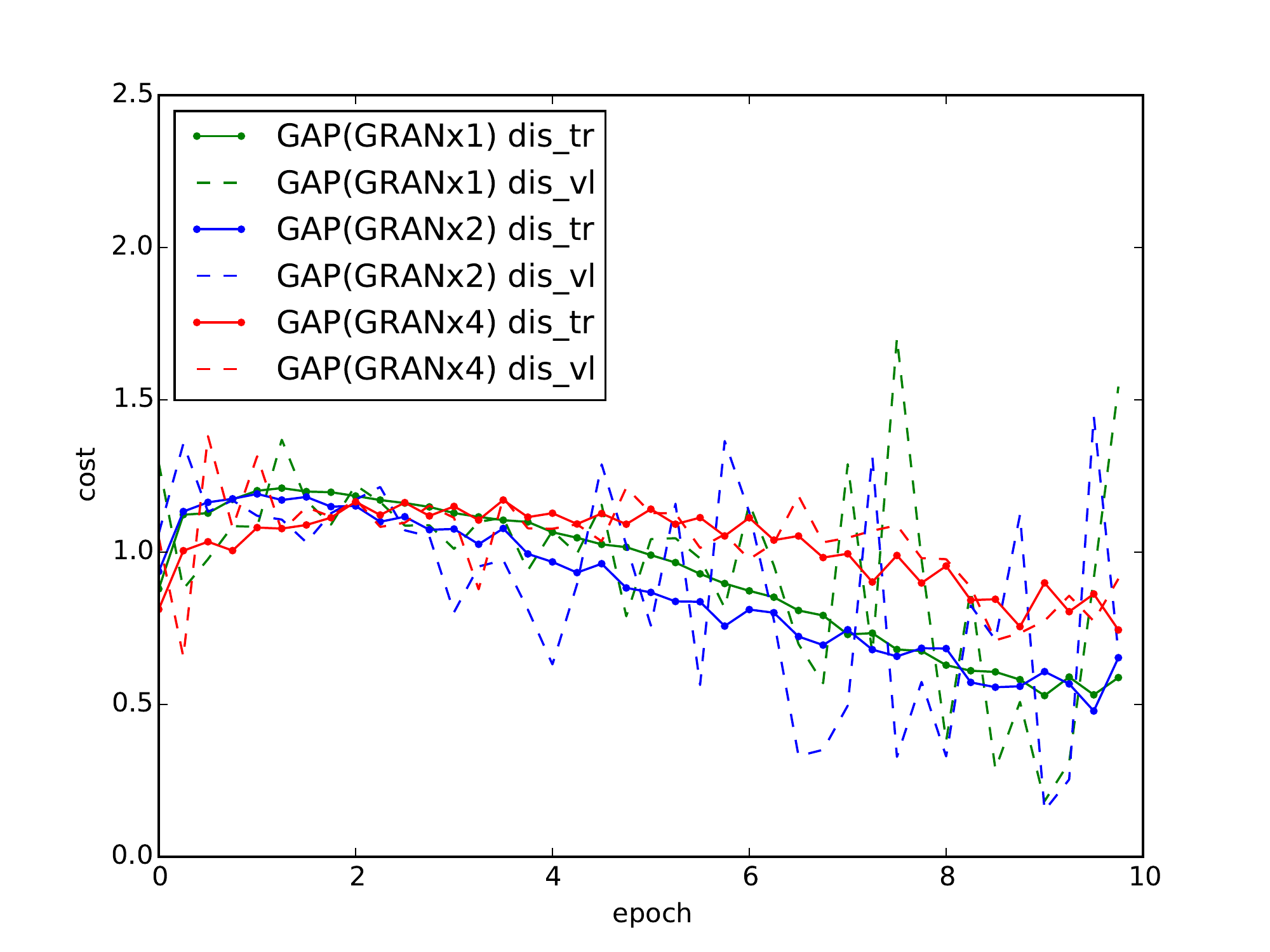}
        \vspace{-0.2cm}
        \subcaption{swapping frequency every half epoch.}
    \end{minipage}
    \begin{minipage}{0.495\textwidth}
        \includegraphics[width=\linewidth]{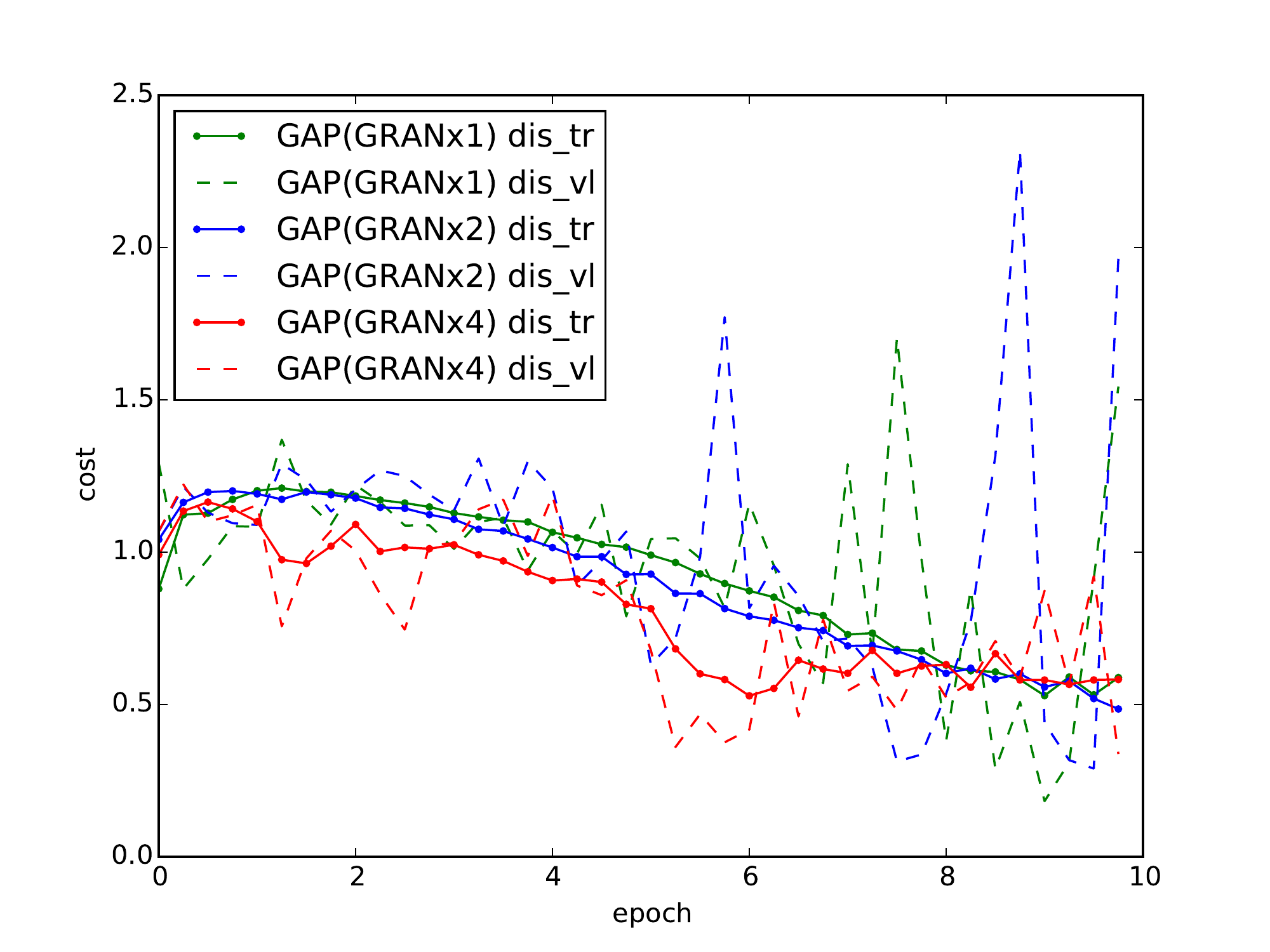}
        \vspace{-0.2cm}
        \subcaption{swapping frequency every epoch.}
    \end{minipage}
    \caption{Averaged GAP(GRAN) learning curves trained on 
        the LSUN Church dataset. As parallelization scales up, 
        the gap between training and validation cost narrows. }
    \label{fig:gap_lsun_learning_curve}
\end{figure}

The supporting figures as in Figure~\ref{fig:gap_cifar10_learning_curve} 
is presented for LSUN dataset in Figure~\ref{fig:gap_lsun_learning_curve}.
There are total of four plots with different swapping frequencies.

\pagebreak

\begin{figure}[htp]
    \begin{minipage}{0.495\textwidth}
        \includegraphics[width=\linewidth]{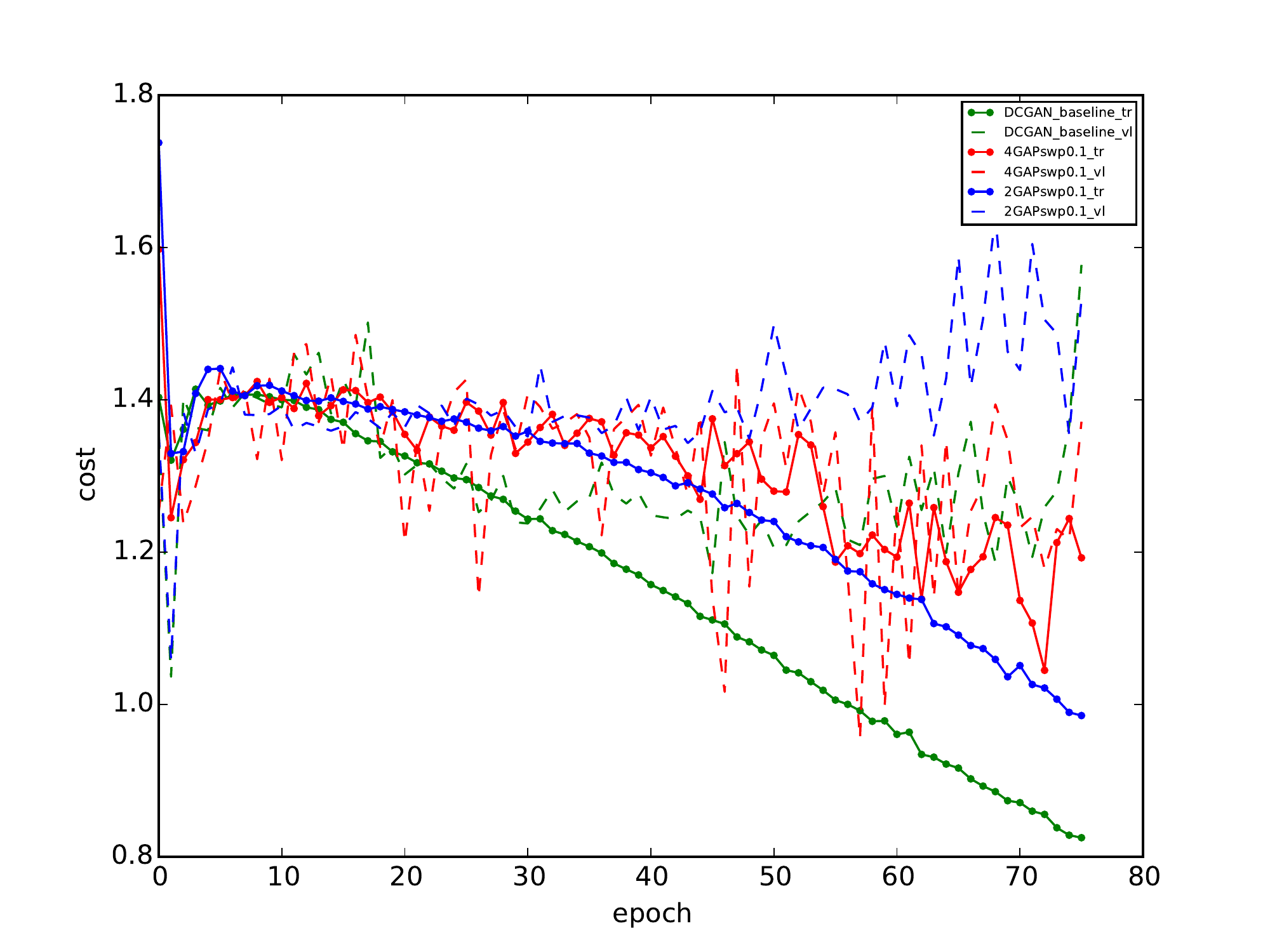}
        \vspace{-0.2cm}
        \subcaption{Swapping frequency every 0.1 epoch.}
    \end{minipage}
    \begin{minipage}{0.495\textwidth}
        \includegraphics[width=\linewidth]{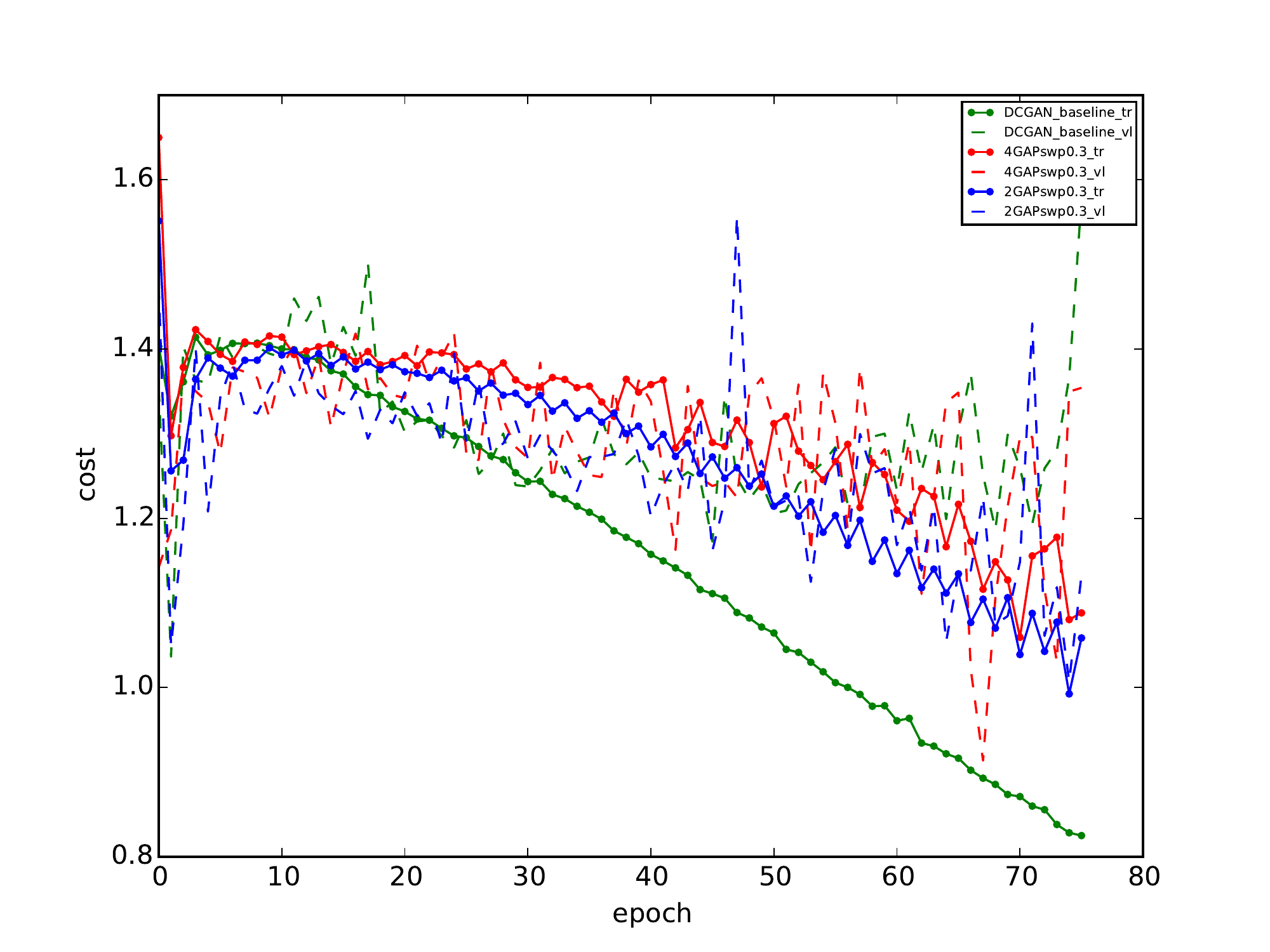}
        \vspace{-0.2cm}
        \subcaption{Swapping frequency every 0.3 epoch.}
    \end{minipage}\\
    \begin{minipage}{0.495\textwidth}
        \includegraphics[width=\linewidth]{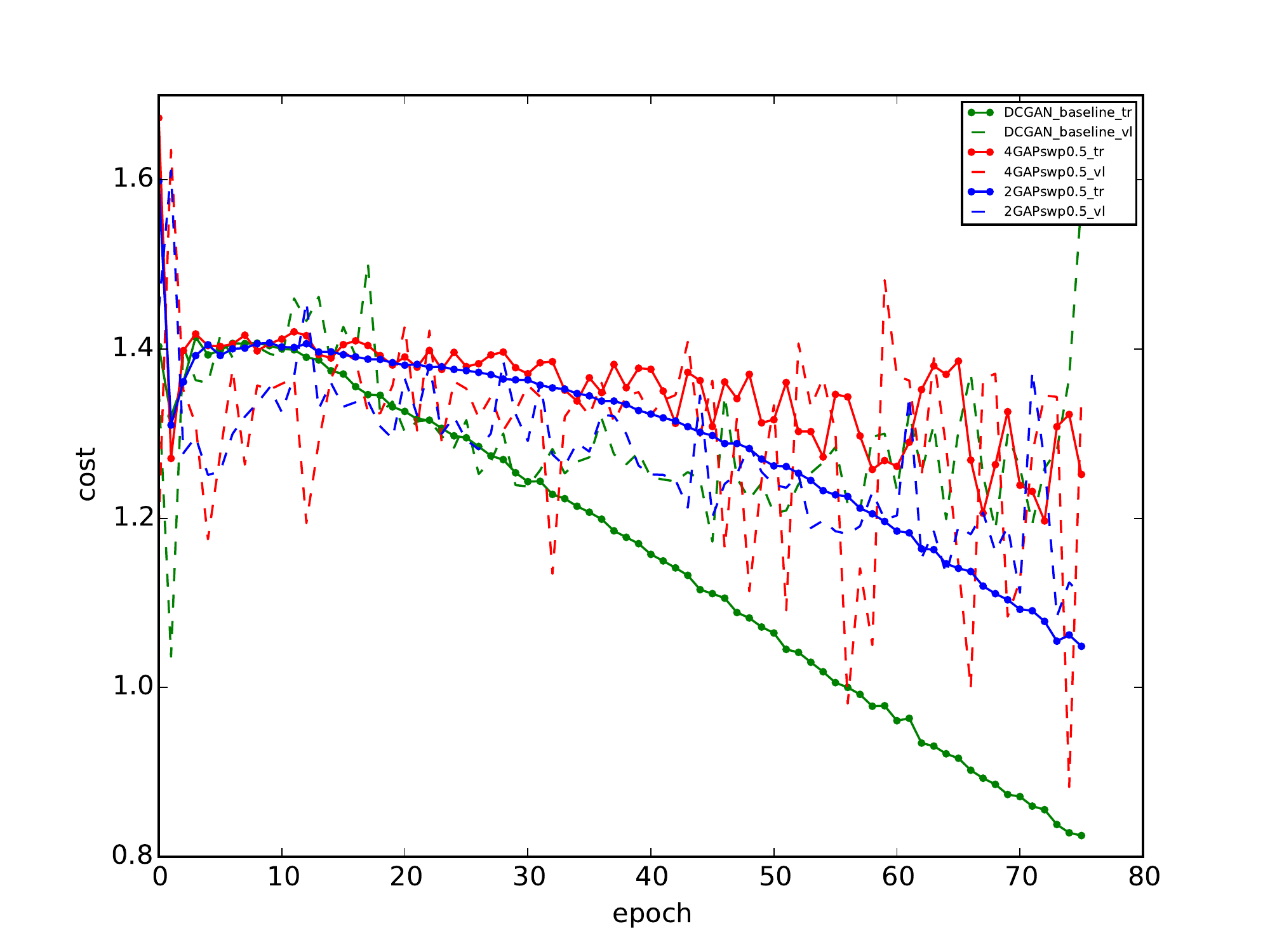}
        \vspace{-0.2cm}
        \subcaption{Swapping frequency every half epoch.}
    \end{minipage}
    \begin{minipage}{0.495\textwidth}
        \includegraphics[width=\linewidth]{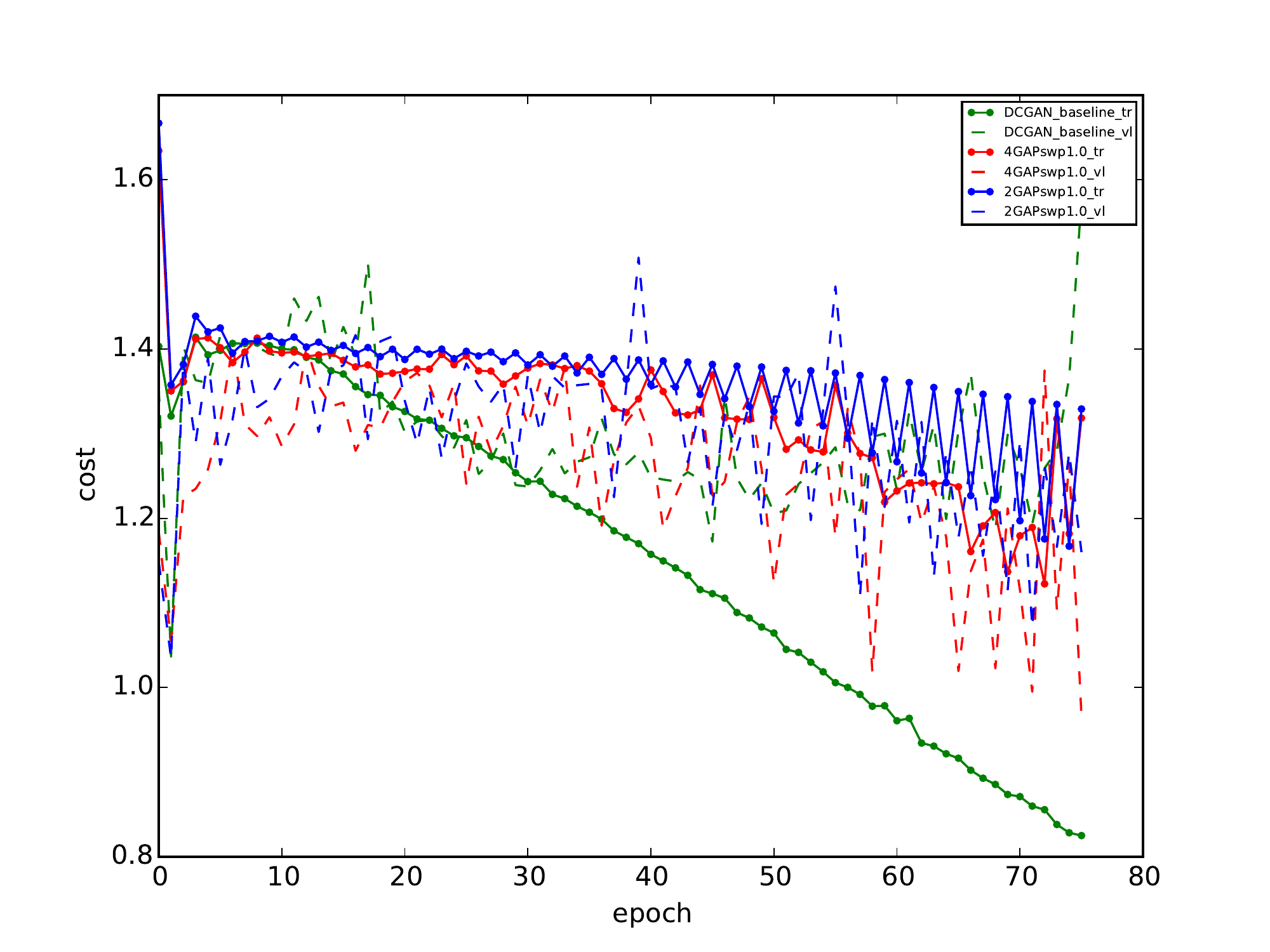}
        \vspace{-0.2cm}
        \subcaption{Swapping frequency every epoch.}
    \end{minipage}
    \caption{Each GAP model represented by the learning curve of a
      single DCGAN within the GAP(DCGAN) trained on CIFAR-10. This
      demonstrates that the observed behaviour of reducing the spread
      between training and validation cost is not simply an effect of
      averaging.}
    \label{fig:lc_swap_rate_single}
\end{figure}

Figure~\ref{fig:lc_swap_rate_single} presents
an instance of an individual learning curve in the case when multiple
GANs are trained under GAP. The difference from
from Figure~\ref{fig:gap_cifar10_learning_curve}
and Figure~\ref{fig:gap_lsun_learning_curve} is that GAP curves are
represented by the learning curve of a single GAN within GAP rather
than an average. Fortunately, the behaviour remains the same, where 
the spread between training and validation cost decreases 
as parallelization scales up (i.e.~more models in a GAP).

\begin{figure}[htp]
    \begin{minipage}{0.245\textwidth}
        \includegraphics[width=\linewidth]{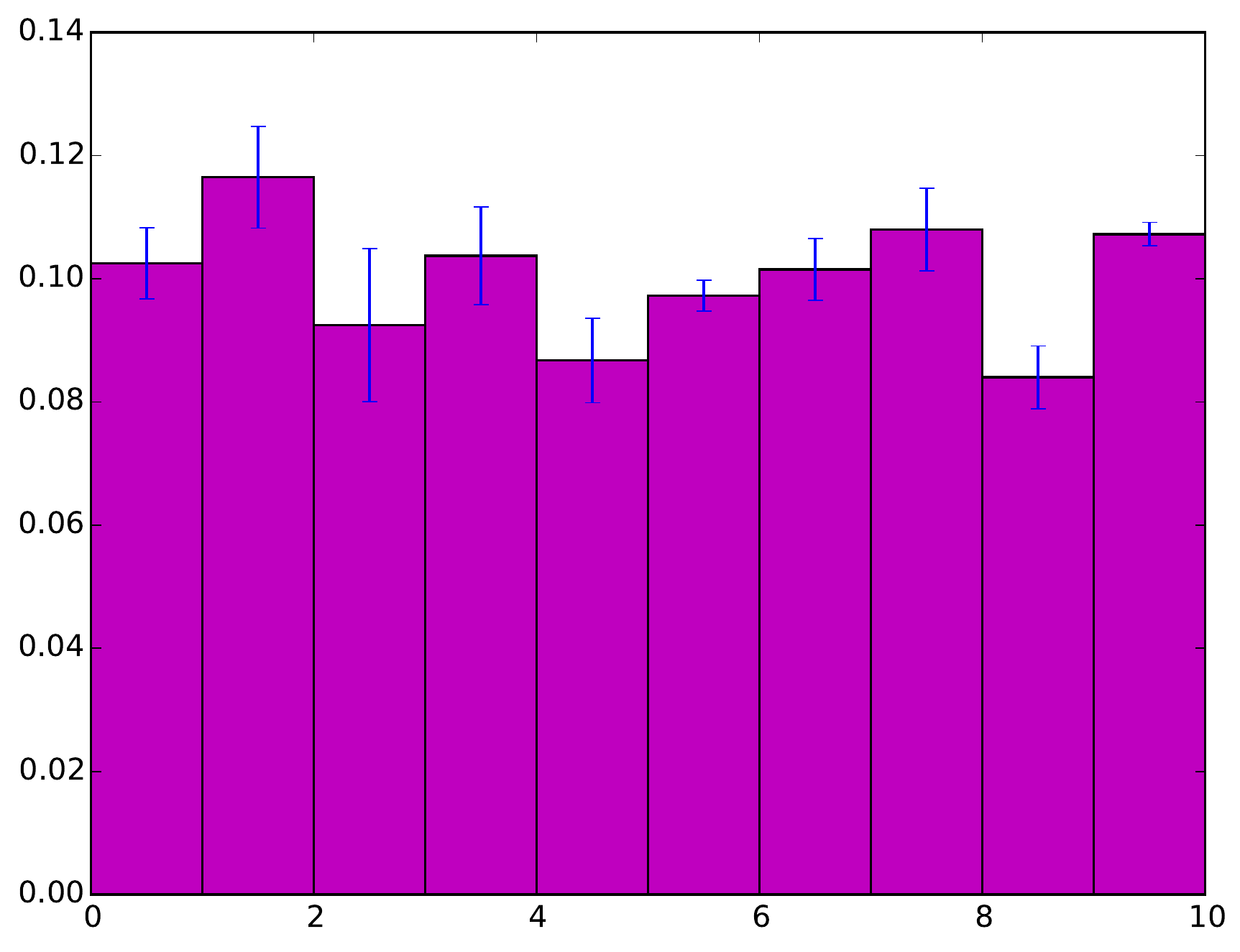}
        \vspace{-0.2cm}
        \subcaption{DCGAN}
    \end{minipage}
    \begin{minipage}{0.245\textwidth}
        \includegraphics[width=\linewidth]{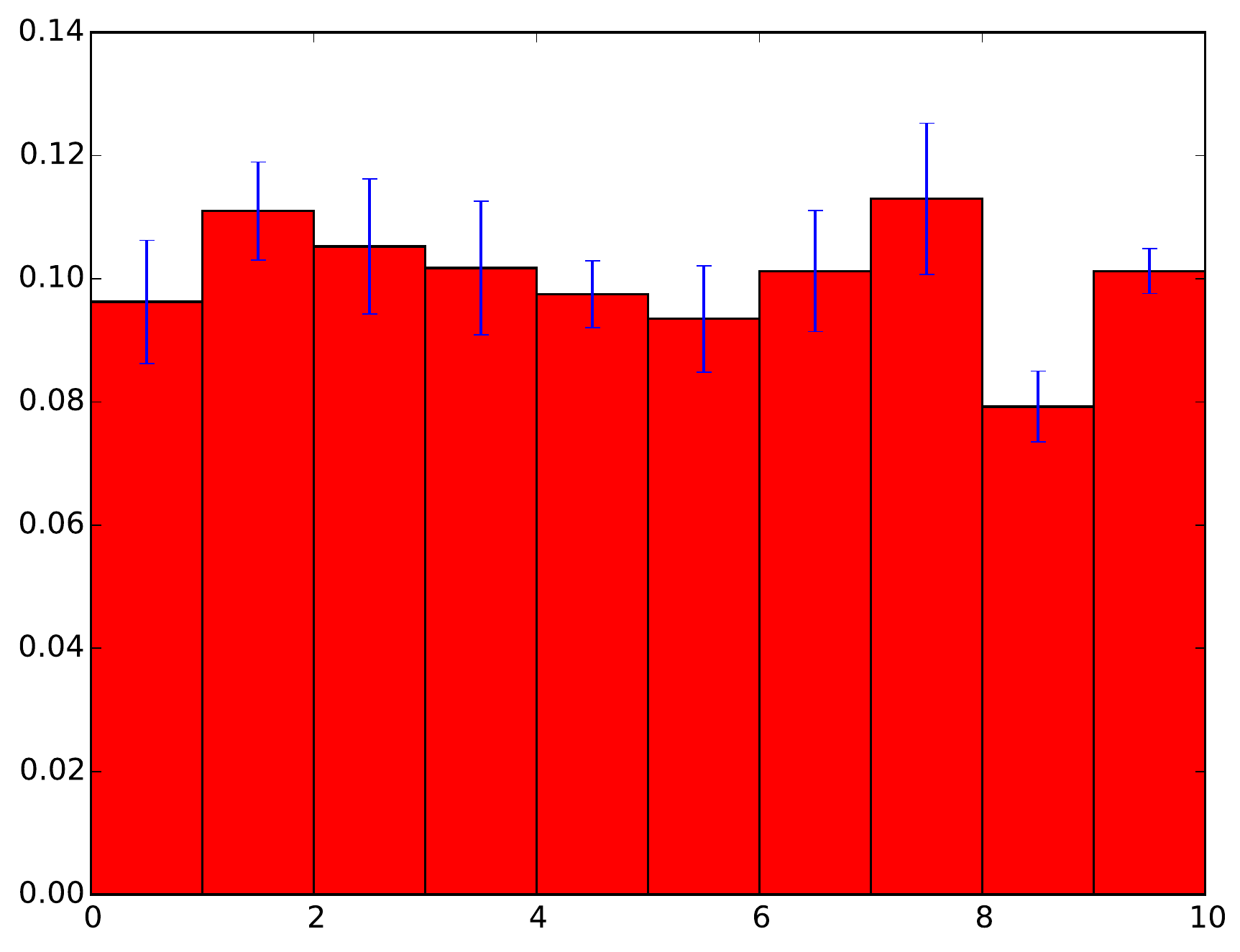}
        \vspace{-0.2cm}
        \subcaption{GAP(DCGANx4)}
    \end{minipage}
    \begin{minipage}{0.245\textwidth}
        \includegraphics[width=\linewidth]{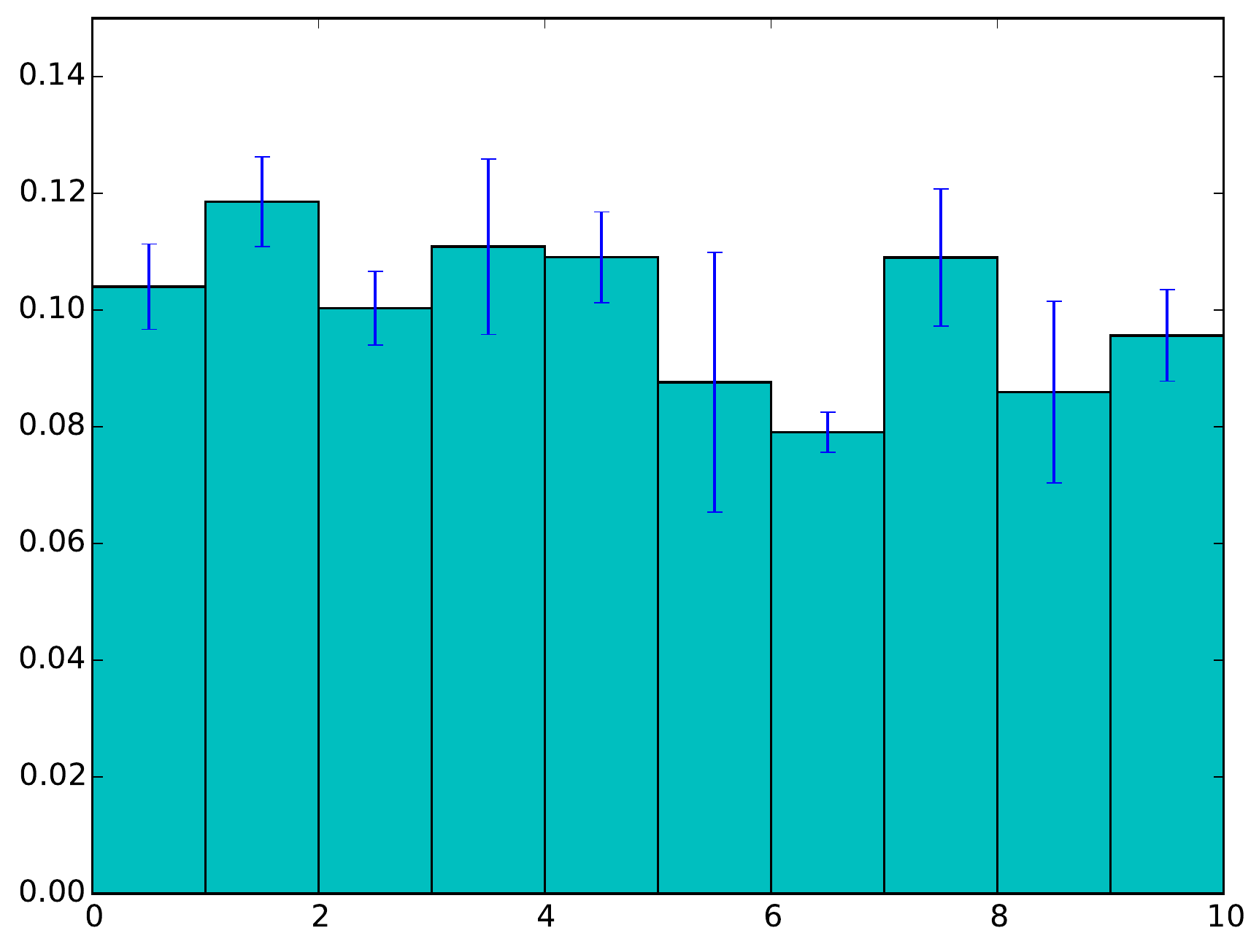}
        \vspace{-0.2cm}
        \subcaption{GRAN}
    \end{minipage}
    \begin{minipage}{0.245\textwidth}
        \includegraphics[width=\linewidth]{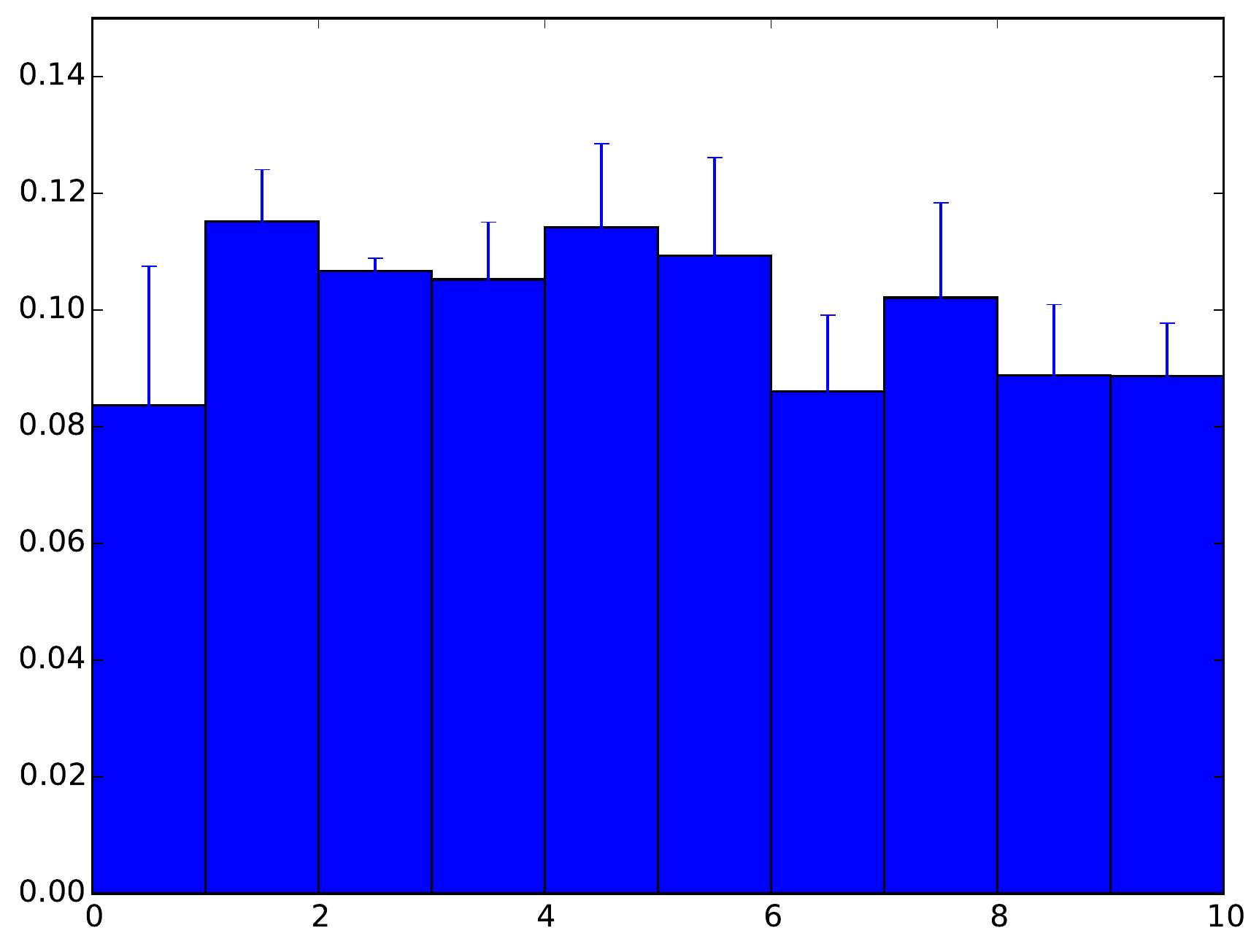}
        \vspace{-0.2cm}
        \subcaption{GAP(GRANx4)}
    \end{minipage}
    \caption{The distribution of the predicted class labels of samples
      from various models
      made by a separately trained logistic regression.}
    \label{fig:hist_mnist_dist}
\end{figure}

\pagebreak We observed the distribution of class predictions on
samples from each model in order to check how closely they match the
training set distribution (which is uniform for MNIST).  We trained a
simple logistic regression on MNIST that resulted in a $\simeq$99\%
test accuracy rate.  The histogram of the predicted classes is
provided in Figure~\ref{fig:hist_mnist_dist}.  We looked at the
exponentiated expected KL divergence between the predicted
distribution and the (uniform) prior distribution, also known as the
``Inception Score'' \citep{Salimans2016}.  The results are shown in
Table~\ref{tab:inception_score}.
\begin{table}[htp]
    \centering
    \caption{Inception Score of GAP}
    \label{tab:inception_score}
    \begin{small}
    \begin{sc}
    \begin{tabular}{c|c|c|c|c}
    \hline
        Models          & DCGAN           & GAP$_{D4}$       & GRAN             & GAP$_{G4}$      \\\hline
        MNIST & $6.36 \pm 0.09$ & $6.592 \pm 0.08$ & $6.77 \pm 0.08$  & $6.85 \pm 0.24$ \\
        %        Score  & 9.26 & 5.29  & 5.54       & 5.14 & 5.43     &  \\
    \hline 
    \end{tabular}
    \end{sc}
    \end{small}
\end{table}
\vspace{-10cm}

\begin{figure}[htp]
    \centering
    \includegraphics[width=\linewidth]{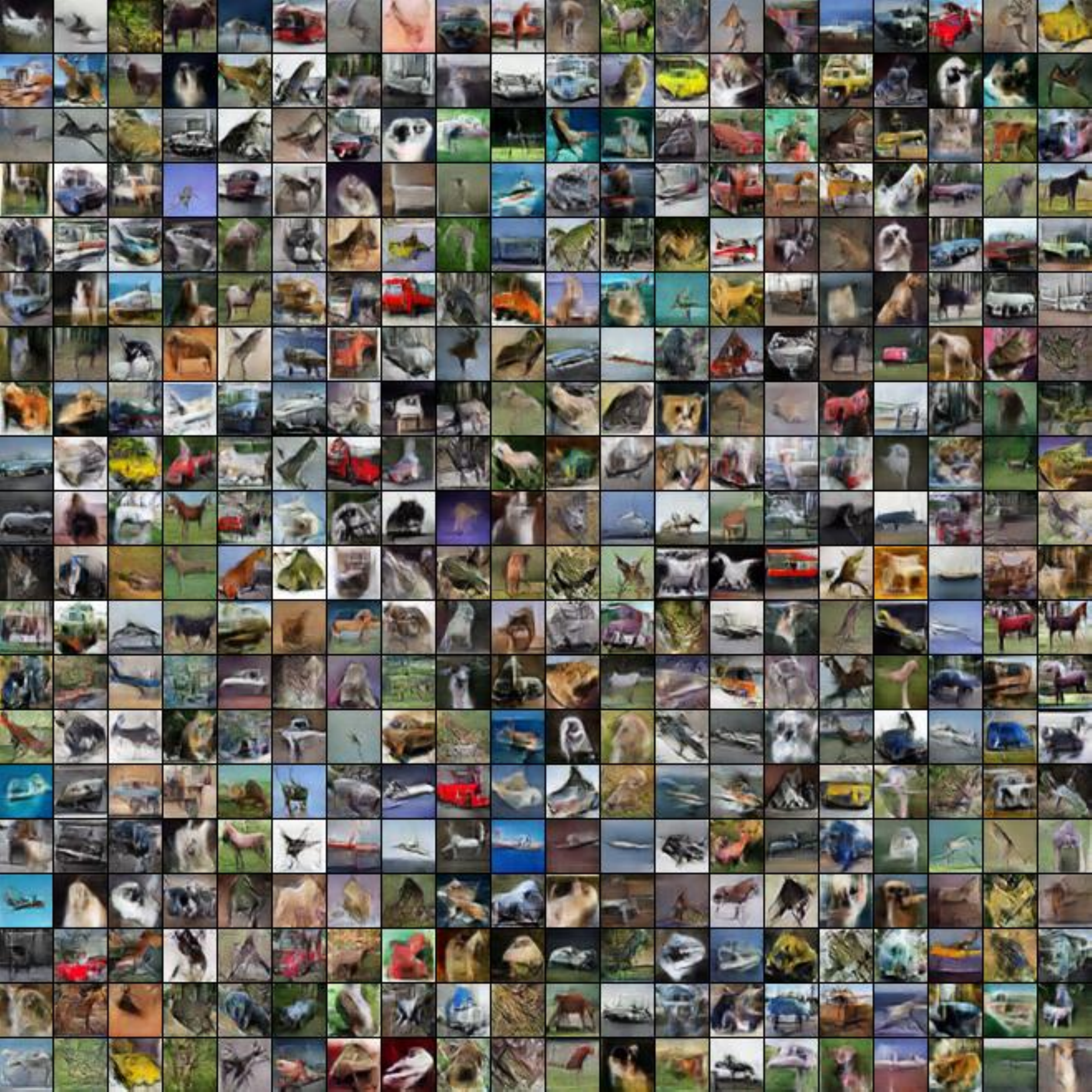}
    \caption{CIFAR-10 samples generated by GAP(DCGANx4). Best viewed in colour.}
    \label{fig:gap_cifar10_samples}
\end{figure}

\begin{figure}[htp]
    \centering
    \includegraphics[width=\linewidth]{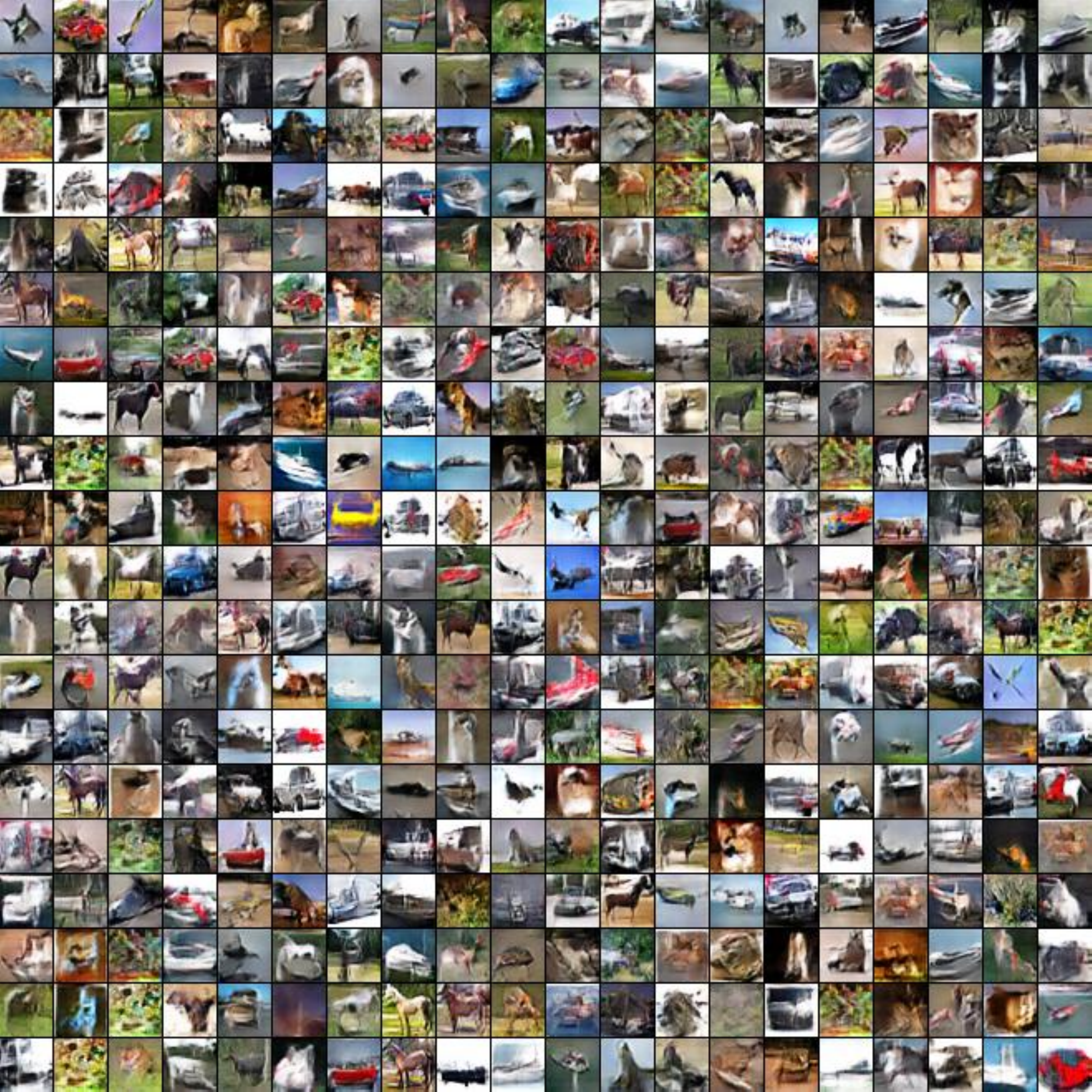}
    \caption{CIFAR-10 samples generated by GAP(GRANx4). Best viewed in colour.}
    \label{fig:gap_cifar10_samples}
\end{figure}

\begin{figure}[htp]
    \centering
    \includegraphics[width=\linewidth]{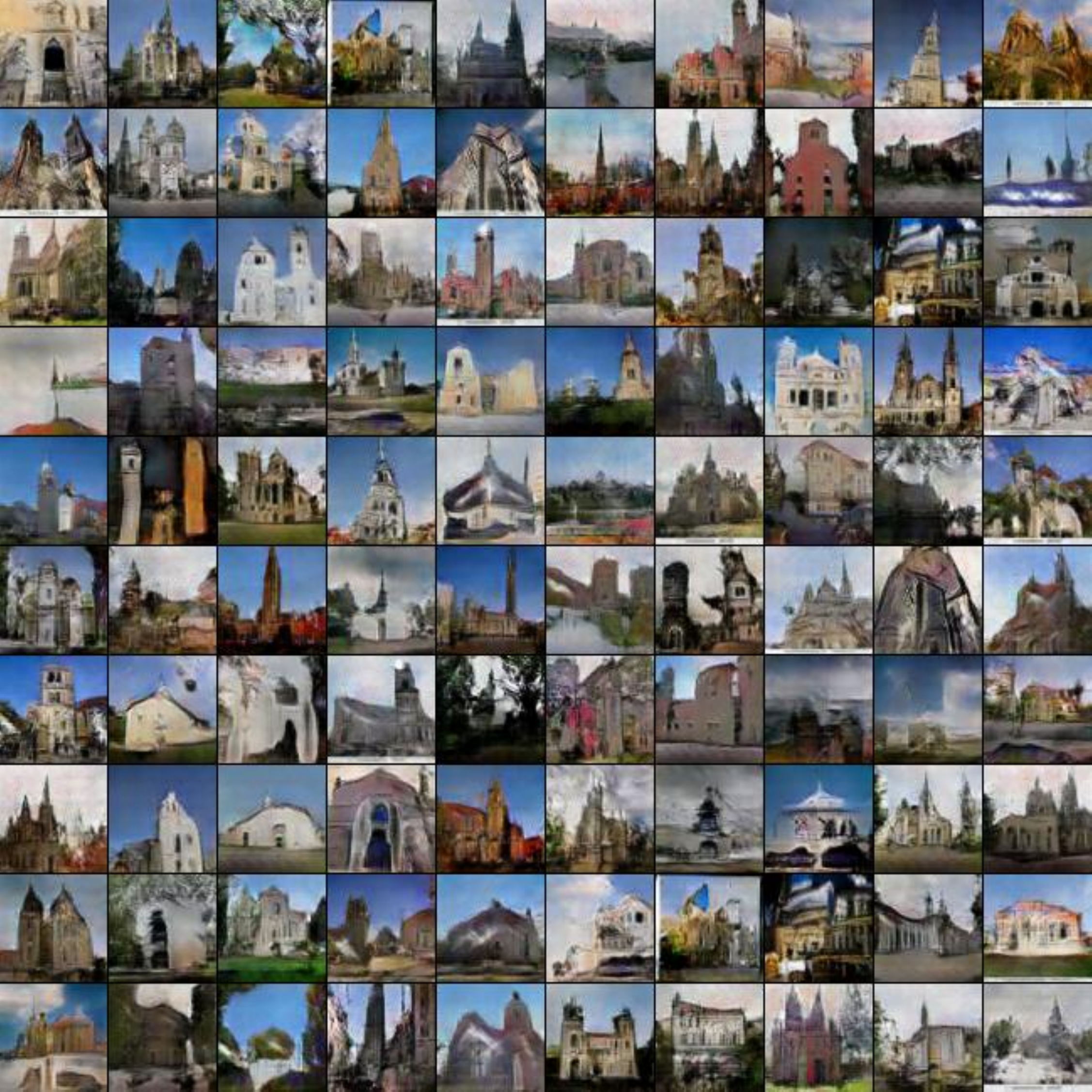}
    \caption{LSUN Church samples generated by GAP(DCGANx4) at 0.3 swapping frequency.
            Best viewed in colour.}
    \label{fig:gap_dcgan_lsun10_samples}
\end{figure}

\begin{figure}[htp]
    \centering
    \includegraphics[width=\linewidth]{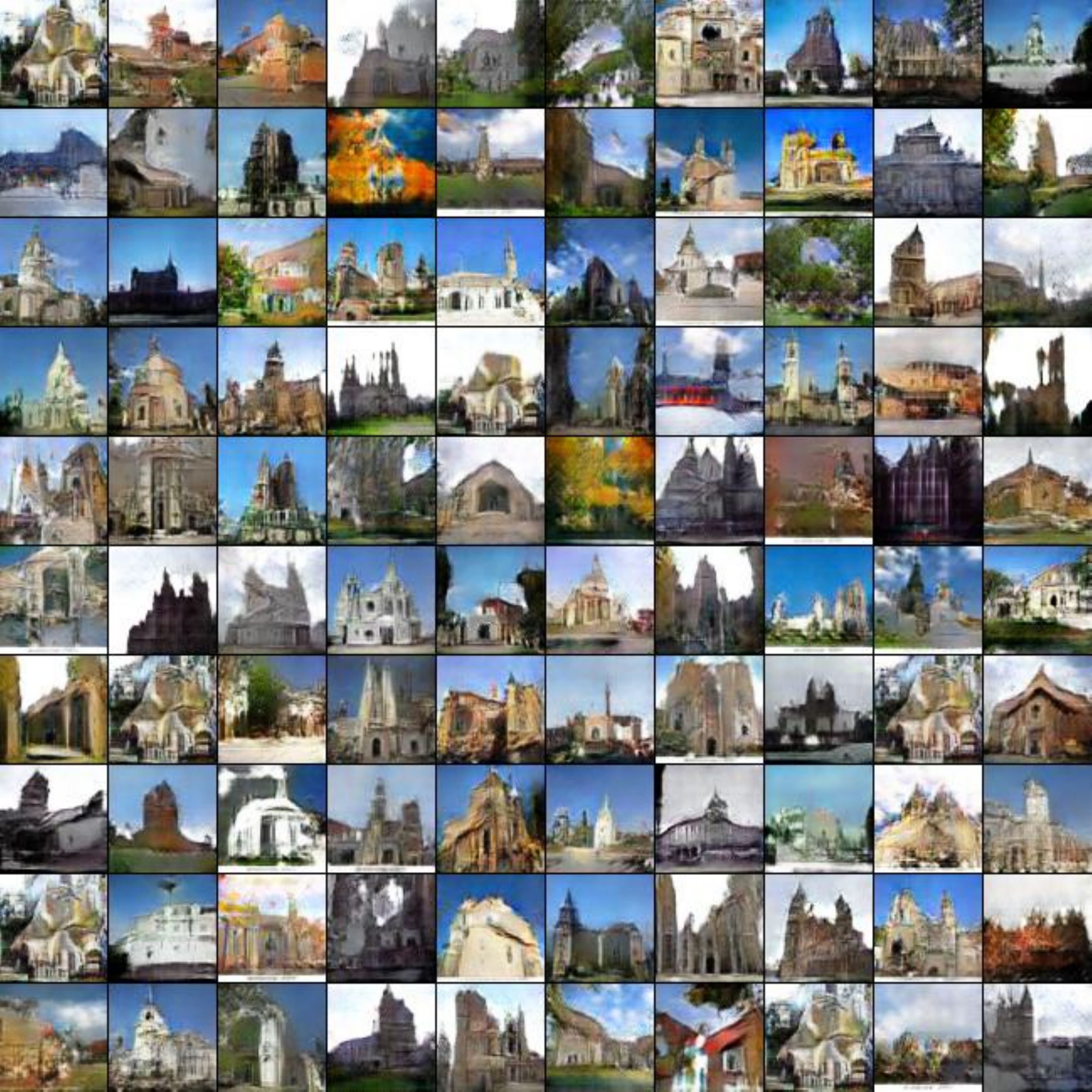}
    \caption{LSUN Church samples generated by GAP(GRANx4) at 0.5 swapping frequency.
            Best viewed in colour.}
    \label{fig:gap_gran_lsun10_samples}
\end{figure}

\pagebreak
%\begin{figure}[htp]
%    \centering
%    \begin{minipage}{0.495\textwidth}
%        \includegraphics[width=\linewidth]{./figs/gap[dcgan4]_samples.pdf}
%    \vspace{-0.2cm}
%    %\subcaption{Samples from one of the DCGANs trained using GAP$_{C4}$}
%    %\label{fig:dcganx4}
%    \end{minipage}
%    \caption{Failure case of capturing various modes: CIFAR-10 samples trained by GAP(DCGANx4).}
%    \label{tab:GAM2}
%\end{figure}
%Here is an example of a failure case in training. Although the samples 
%look like real images,  the samples from one of the DCGANs trained by 
%GAP are still dominated by dogs.
%Nevertheless, this is one of DCGAN among four DCGANs trained.
%This implies that some of the DCGANs trained by GAP can still drop modes. 

\subsubsection{Fine-tuning GANs using GAP}
\begin{figure}[htp]
    \centering
    \begin{minipage}{0.495\textwidth}
        \includegraphics[width=\linewidth]{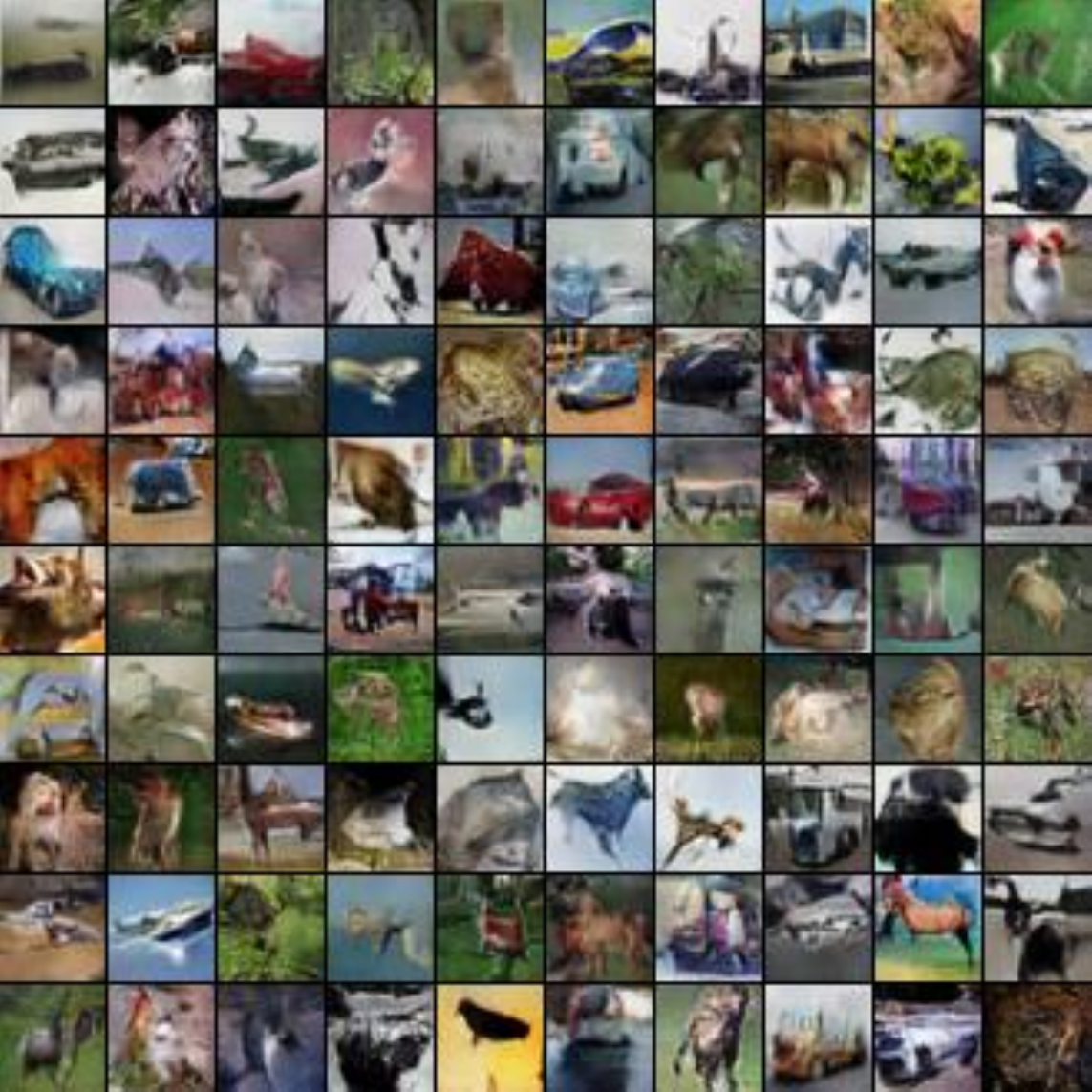}
        \vspace{-0.2cm}
        \subcaption{Samples from one of the DCGANs trained using GAP$_{C4}$}
        \label{fig:comb_dcganx4}
    \end{minipage}
    \begin{minipage}{0.49\textwidth}
        \includegraphics[width=\linewidth]{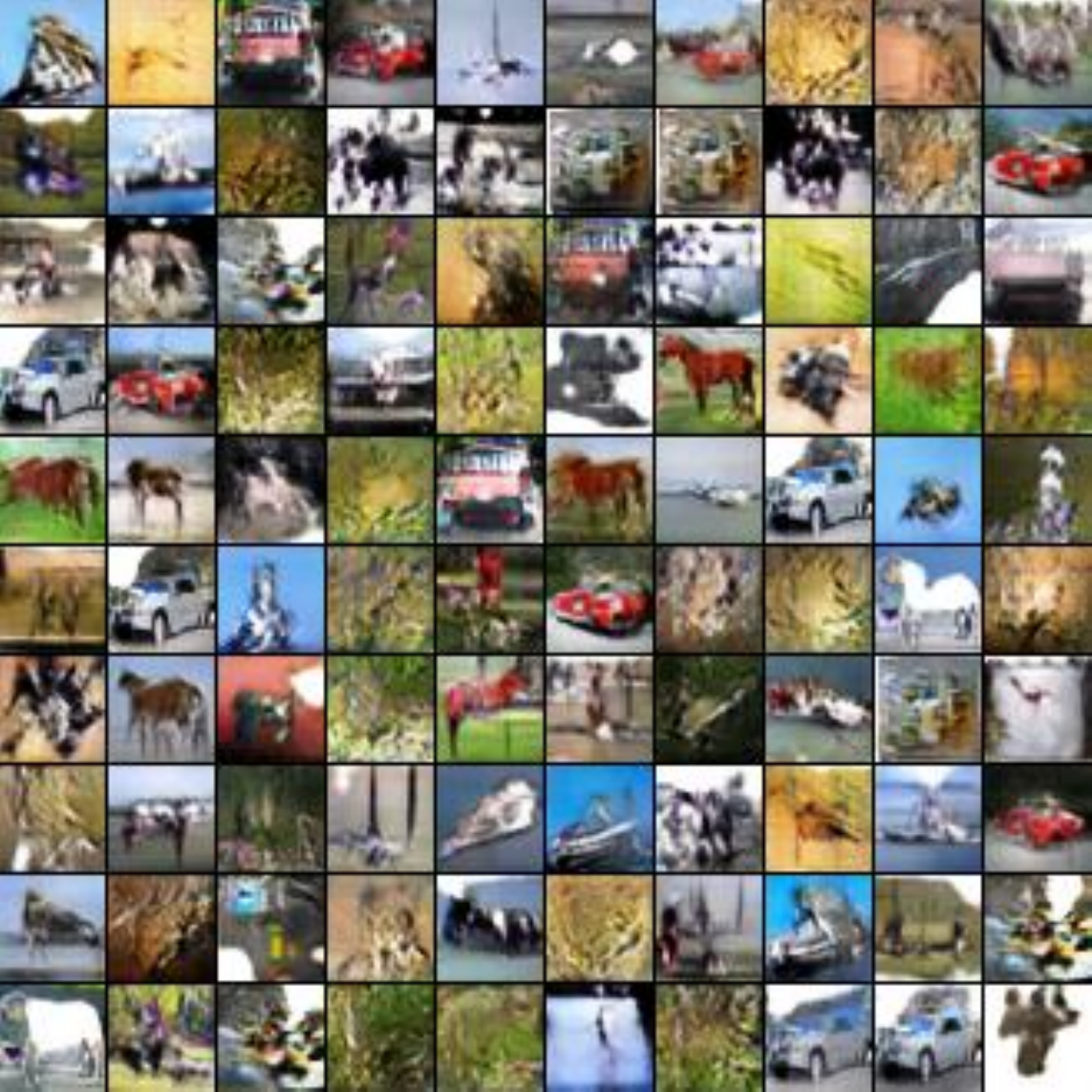}
        \vspace{-0.2cm}
        \subcaption{Samples from one of the GRANs trained using GAP$_{C4}$}
        \label{fig:comb_granx4}
    \end{minipage}\\
    \caption{CIFAR-10 samples trained by GAP(DCGANx2, GRANx2). }
    \label{fig:comb_gap_cifar10_samples}
    \begin{minipage}{0.49\textwidth}
        \includegraphics[width=\linewidth]{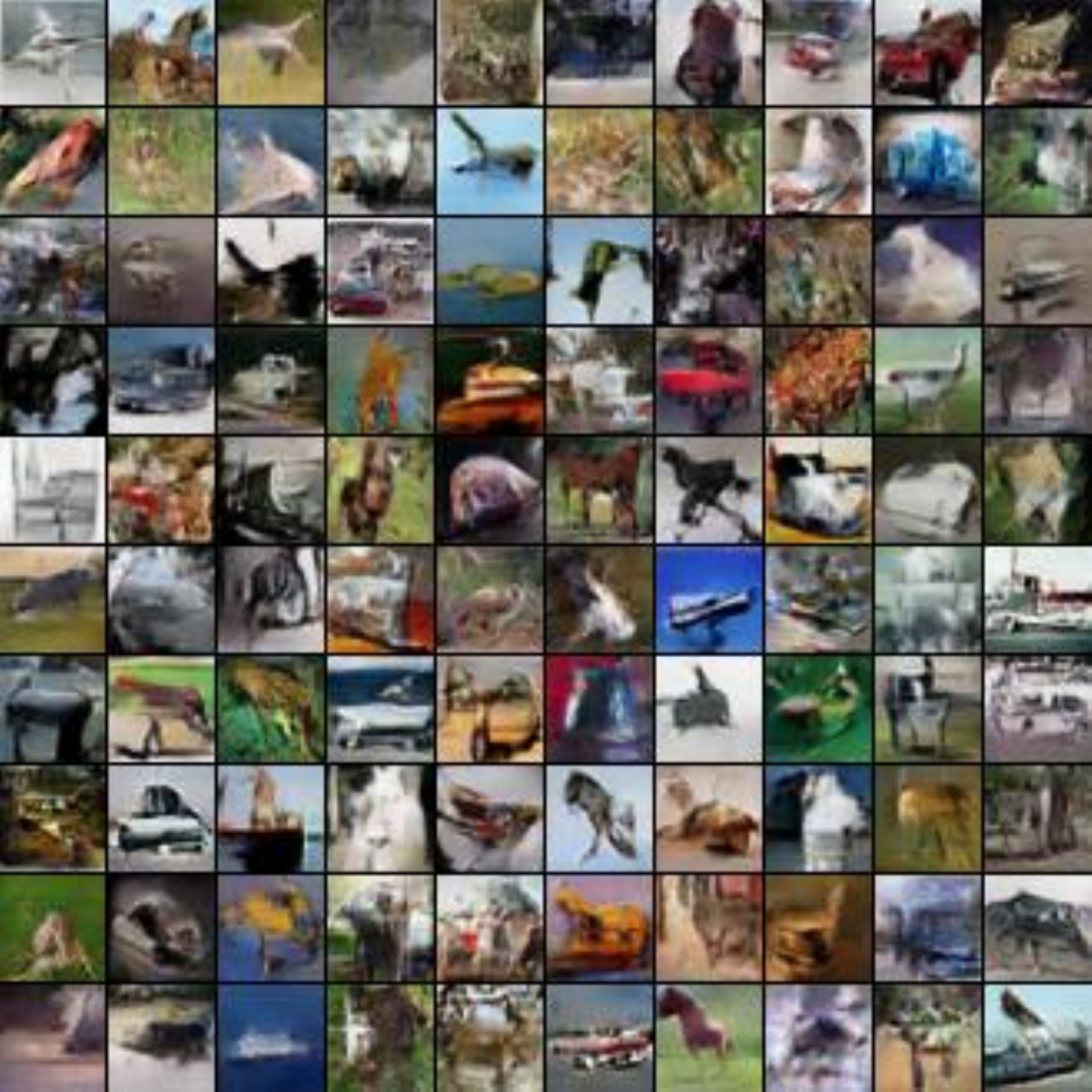}
        \subcaption{Samples from one of the fine-tuned DCGAN using GAP$_{C4}$}
        \label{fig:fcomb_dcganx4}
    \end{minipage}
    \begin{minipage}{0.49\textwidth}
        \includegraphics[width=\linewidth]{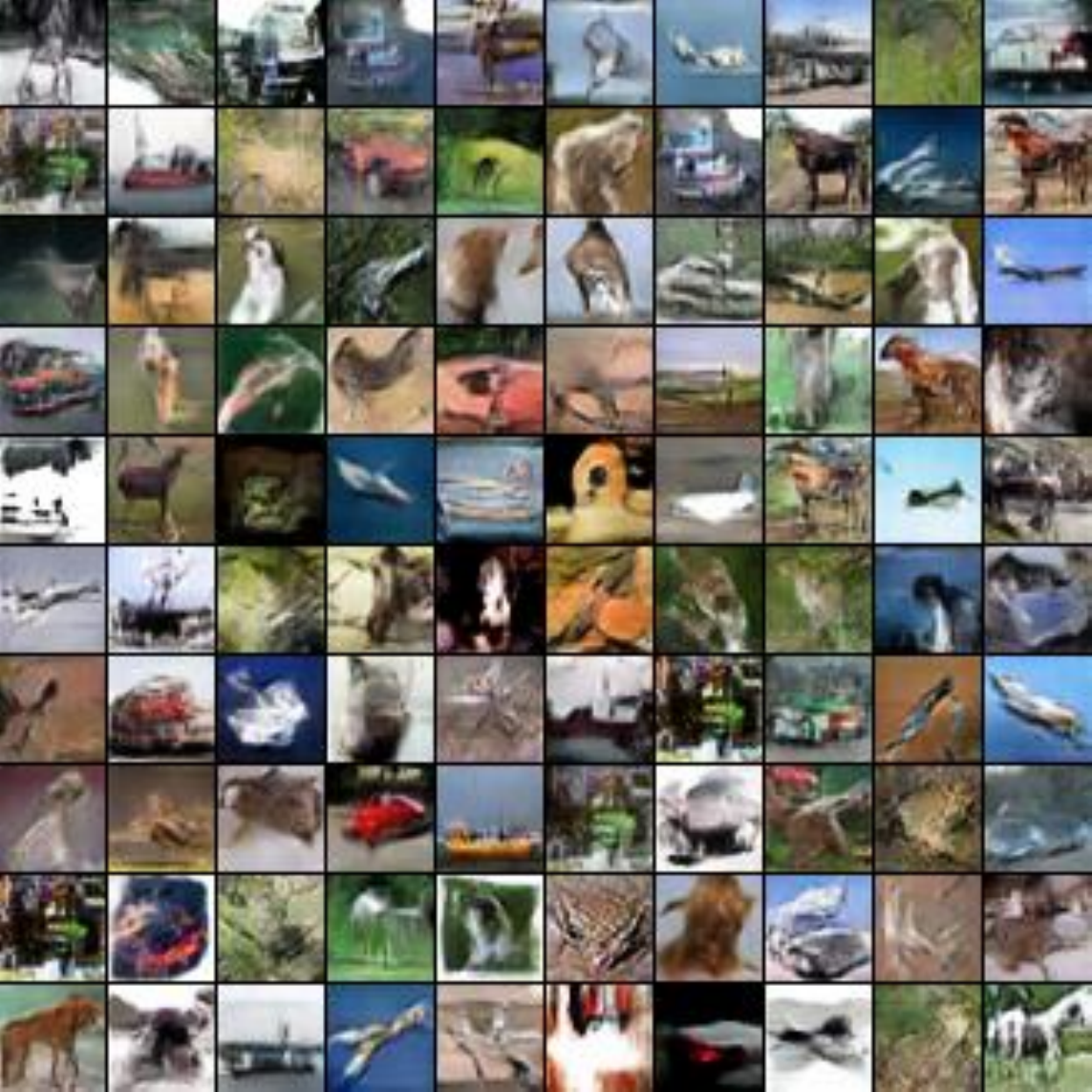}
        \subcaption{Samples from one of the fine-tuned GRAN using GAP$_{C4}$}
        \label{fig:fcomb_granx4}
    \end{minipage}
    \caption{CIFAR-10 samples trained by GAP(DCGANx2, GRANx2).}
    \label{fig:fcomb_gap_cifar10_samples}
\end{figure}

We also tried fine-tuning individually-trained GANs using GAP, which
we denote as GAP$_{F4}$.  GAP$_{F4}$ consists of two trained DCGANs
and two trained GRANs.  They are then fine-tuned using GAP for five
epochs.  Samples from the fine-tuned models are shown in
Figure~\ref{fig:fcomb_gap_cifar10_samples}.
%However, GAP$_{F4}$ was not succesf
%We compared these two models against GAP(DCGAN) and GAP(GRAN)

\pagebreak

\end{document}